\documentclass[10pt,twocolumn,letterpaper]{article}

\usepackage[pagenumbers]{cvpr} 
\usepackage[accsupp]{axessibility} 
\usepackage{graphicx}
\usepackage{amsmath}
\usepackage{amssymb}
\usepackage{booktabs}

\usepackage{mathtools}

\usepackage[pagebackref,breaklinks,colorlinks]{hyperref}

\usepackage[capitalize]{cleveref}
\crefname{section}{Sec.}{Secs.}
\Crefname{section}{Section}{Sections}
\Crefname{table}{Table}{Tables}
\crefname{table}{Tab.}{Tabs.}

\def\cvprPaperID{8} 
\def\confName{CVPR}
\def\confYear{2022}

\begin{document}

\title{High-Resolution UAV Image Generation for Sorghum Panicle Detection}

\author{Enyu Cai, Zhankun Luo, Sriram Baireddy, Jiaqi Guo, Changye Yang, Edward J. Delp\\
School of Electrical and Computer Engineering \\
Purdue University \\
West Lafayette, Indiana\\
{\tt\small \{cenyu, luo333, sbairedd, guo498, yang853, ace\}@ecn.purdue.edu}
}
\maketitle

\begin{abstract}
    The number of panicles (or heads) of  Sorghum plants is an important phenotypic trait for plant development and grain yield estimation.
    The use of Unmanned Aerial Vehicles (UAVs) enables the capability of collecting and analyzing Sorghum images on a large scale.
    Deep learning can provide methods for estimating phenotypic traits from UAV images but requires a large amount of labeled data.
    The lack of training data due to the labor-intensive ground truthing of UAV images causes a major bottleneck in developing methods for Sorghum panicle detection and counting.
    In this paper, we present an approach that uses synthetic training images from generative adversarial networks (GANs) for data augmentation to enhance the performance of Sorghum panicle detection and counting.
    Our method can generate synthetic high-resolution UAV RGB images with panicle labels by using image-to-image translation GANs with a limited ground truth dataset of real UAV RGB images.
    The results show the improvements in panicle detection and counting using our data augmentation approach.
\end{abstract}

\section{Introduction}
\label{sec:intro}

Crop-based biofuels, an environmentally sustainable energy resource derived from plant matter, can help reduce greenhouse gas emissions and dependency on exhaustible resources\cite{bio_fuel,sweet_sorghum}.
In recent studies, Sorghum (\textit{Sorghum bicolor} (L.) Moench) has been found to be a very good potential biofuel \cite{sweet_sorghum,rao_2015, velmurugan_2020}.
Biofuel from Sorghum is less of a concern than biofuel from main food crops such as corn and sugar cane since the production of biofuel from these food crops could lead to supply shortages in food production\cite{rao_2015}.
\begin{figure}[t]
  \centering
  \begin{subfigure}{0.45\textwidth}
    \centering\includegraphics[width=0.61\linewidth]{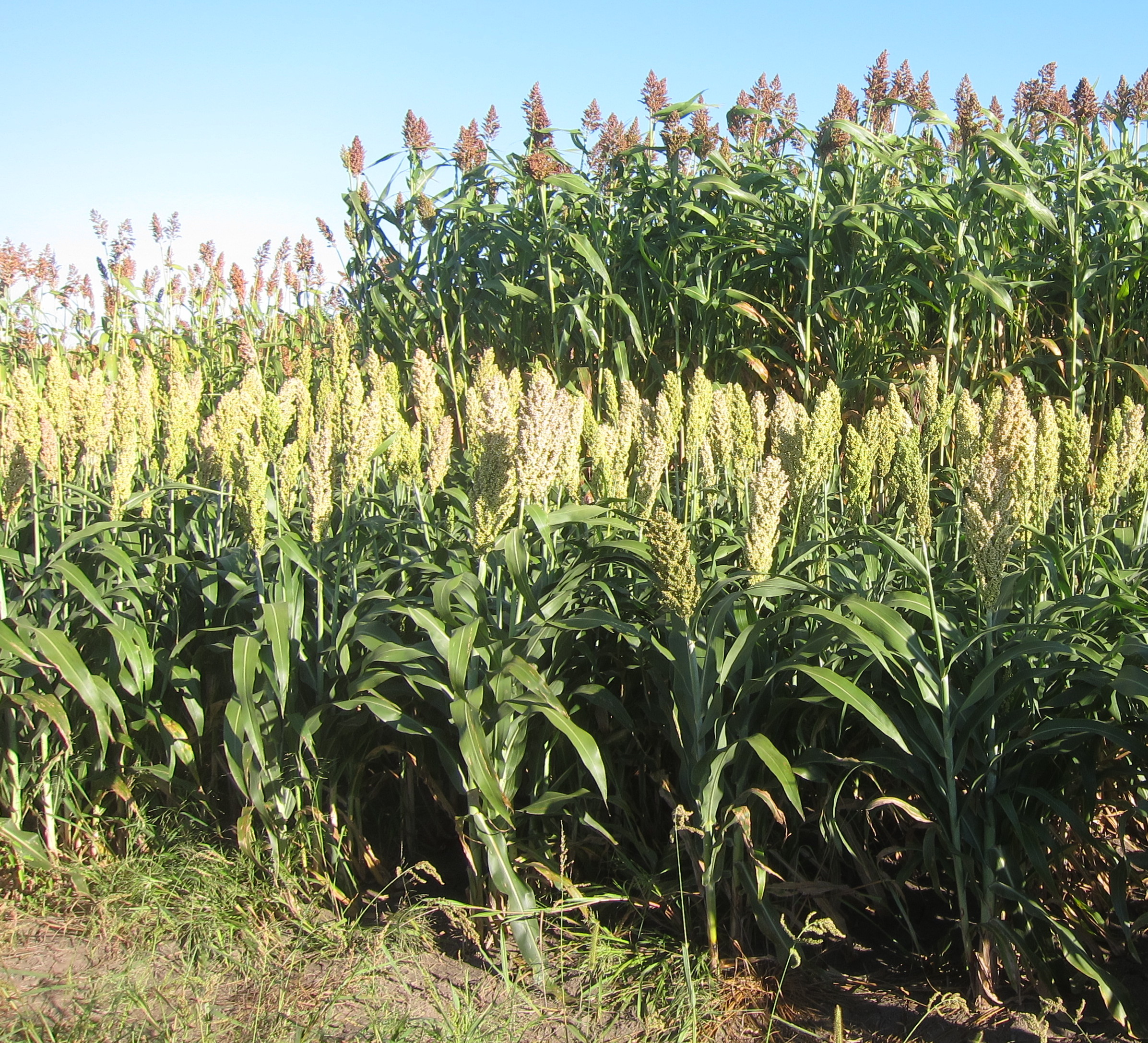}
    \caption{A Sorghum field (from \cite{javithesis}). The panicle is the cluster structure at the top of the Sorghum plants.}
  \end{subfigure}
	\begin{subfigure}{0.45\textwidth}
        \centering\includegraphics[width=0.59\linewidth]{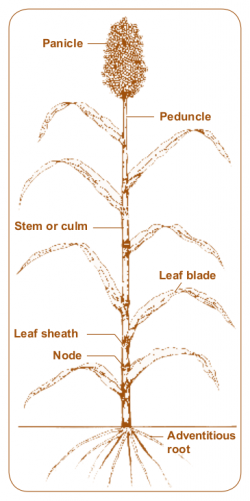}
        \caption{Anatomy of Sorghum (from \cite{sorghum_structure}).}
      \end{subfigure}
  \caption{Sorghum panicles.}
  \label{fig:panicle}
 \end{figure}
A Sorghum panicle is a cluster of flowers at the top of the plant as shown in \Cref{fig:panicle}.
The ``panicle'' (or ``head'') of Sorghum is one of the most important features of the plant related to grain yield, plant growth, and management\cite{ezeaku_2006,pend_1994}.
Detecting and counting the panicles can provide information such as flowering dates to help plant breeders select the proper type of crops\cite{cai_2021,ezeaku_2006,pend_1994,lin_2020}.

Plant phenotyping is the methodology of linking a plant's observable characteristics to genetic information by estimating the phenotypic traits of plants such as height, shape, quantity, and color\cite{walter_2015}.
Traditional field-based phenotyping usually consists of walking through the field and manually estimating the plant traits using destructive methods which are not suitable on a large scale\cite{addie_paper}.
To address the limitations of traditional phenotyping, modern high-throughput phenotyping\cite{furbank_2011,chapman_2014,khan2018estimation,lootens2016high} uses remote sensing with Unmanned Aerial Vehicles (UAVs) to collect data.
The use of UAVs has dramatically improved phenotyping in 
speed, scale, and quality in a non-destructive manner by flying at low altitudes and collecting data rapidly.
UAVs equipped with RGB sensors provide images with high spatial and temporal resolution for plant trait analysis.
Typically, images acquired by UAVs are orthorectified with accurate location properties 
into orthomosaic\cite{habib} for data processing. This allows plant traits to be accurately measured.

Deep convolutional neural networks (CNN) \cite{alexnet} have achieved outstanding results in image classification, object detection and segmentation\cite{vgg,fasterrcnn,maskrcnn}.
The high accuracy for identifying objects in images with complicated backgrounds makes CNNs attractive for estimating phenotypic traits.
For deep neural networks, a large quantity of training data is often required to prevent overfitting.
Labeling a large amount of training data is time-consuming and tedious due to the high spatial resolution and density of panicles in plant images.
The lack of training data is a major bottleneck that affects the performance of the deep neural network in plant trait estimation.
Data augmentation\cite{data_augmentation} is a technique to reduce overfitting by creating more training data samples such as rotation, flip, and crop on the existing dataset without additional ground truth.
Adding synthetic images has become popular for data augmentation since the introduction of generative adversarial networks (GANs)\cite{gan}, which can generate realistic images.
GAN-based methods have a positive impact for enhancing the performance for classification tasks for plant images \cite{bi_2020, madsen_2019}.
Most GAN-based approaches focus on classification by generating images that contain a single instance of the plant.
In plant trait estimation such as panicle detection, there are multiple instances of plants in an image, so the ability to generate images containing multiple objects with labels is necessary.

\begin{figure}[t]
  \centering
  \begin{subfigure}{0.45\textwidth}
    \centering\includegraphics[width=0.9\linewidth]{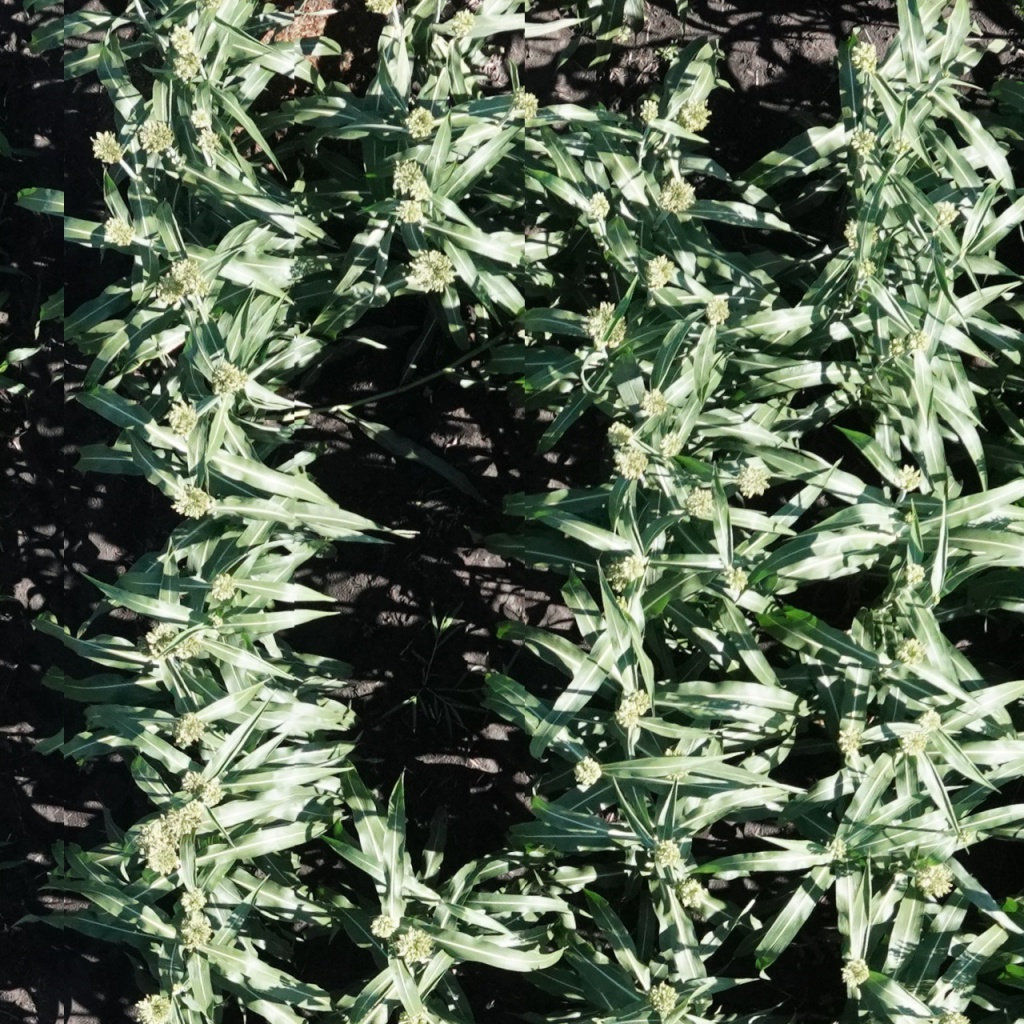}
    \caption{A real UAV RGB image of a Sorghum field.}
  \end{subfigure}
	\begin{subfigure}{0.45\textwidth}
        \centering\includegraphics[width=0.9\linewidth]{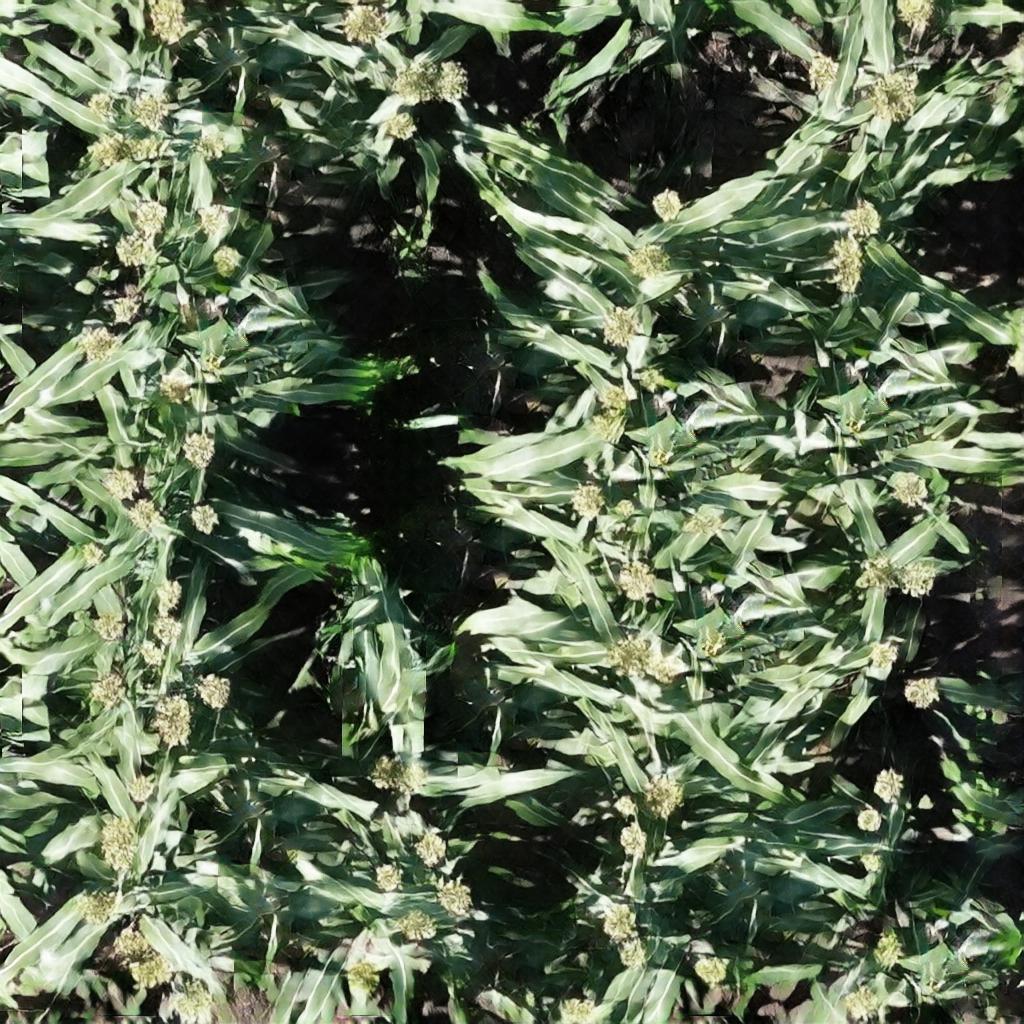}
        \caption{A synthetic UAV RGB image of a Sorghum field generated from our GAN.}
      \end{subfigure}
  \caption{An example of real and synthetic UAV RGB images of a Sorghum field.}
  \label{fig:ex_result}
\end{figure}

In this paper, we propose a method for generating high-resolution synthetic Sorghum UAV RGB images with bounding box labels 
using two image-to-image translation GANs.
We create a label map (or label images) that contains bounding boxes where we want the panicle to be located in the synthetic image generated by the GAN. We shall denote each bounding box in the label map as a "mask".
The GAN then generates a synthetic image with the panicle located as proscribed 
in the label map.
We utilize a small amount of ground truth real images to train the image-to-image translation GAN with the label map.
The synthetic images are combined with real images to train the object detection network.
We compare the performance of the two image-to-image translation GANs.
The results show performance improvement in both panicle detection and counting. 
\Cref{fig:ex_result} shows an example of real and synthetic UAV RGB images of Sorghum.

\section{Related Work}
\label{sec:related}
\textbf{Generative Adversarial Network (GAN).}
The basic idea of a GAN\cite{gan} is to generate images that are indistinguishable from real images.
A GAN consists of two models: the generator model for image generation and the discriminator model for classification of real and synthetic images.
The generator often uses an inside-out autoencoder\cite{autoencoder} structure that takes random noise as input and generates synthetic images.
The discriminator uses a CNN and multilayer perceptron (MLP)\cite{mlp} as a classifier to distinguish the real and synthetic data.
During the training of the GAN, the objective of the discriminator is to maximize the cost value, while the objective of the generator is to minimize the cost value.
Developing a stable training approach is the main challenge in recent GAN research\cite{gan_survey}.
Since the GAN has two models (Generator and Discriminator), it is difficult to train both models simultaneously while ensuring convergence.
There are a variety of GAN variants that focus on improving training stability.
For example, Arjovsky \etal proposed the Wasserstein GAN\cite{wgan} with a new loss function based on the Earth Mover's Distance\cite{em_distance} to enhance training stability.
Radford \etal developed a new model DCGAN\cite{dcgan} using a novel CNN architecture for both the generator and discriminator.
The Conditional GAN (cGAN)\cite{conditional_gan} is a variant of a GAN in which auxiliary inputs, such as labels or text, are added as inputs to both the generator and discriminator so that the output can be controlled by the additional information.
The Conditional GAN has wide applications such as text-to-image translation\cite{text_to_image}, image-to-image translation\cite{image2image_translation}, and style transfer\cite{style_gan}.

\textbf{Data Augmentation Using Synthetic Images.}
Generating synthetic images from GANs for data augmentation is widely used to enhance the performance of deep neural networks for various tasks where there is insufficient training data\cite{data_augmentation}.
The synthetic data is combined with real ground truth image to train a target network.
Most GAN-based data augmentation methods are used to improve  classification that identifies the single object from images.
For example, in \cite{bi_2020}, Bi \etal developed a method using images generated by a Wasserstein GAN variant to improve plant disease classification.
In \cite{madsen_2019}, Madsen \etal improved plant seeds classification by using synthetic plant seed images from GAN.
The Conditional GAN is often used in situations that require additional annotations for the synthetic images, such as data augmentation for detection and segmentation tasks.
For example, in \cite{aerialgan}, Milz \etal proposed a conditional GAN to generate images with ground truth data for aerial object detection and segmentation.
In \cite{sandfort_2019}, Sandfort \etal improved  CT segmentation by using synthetic images from CycleGAN\cite{cyclegan}.

\section{Proposed Approach}
\label{sec:method}
\begin{figure}[t]
    \centering
    \includegraphics[width=0.47\textwidth]{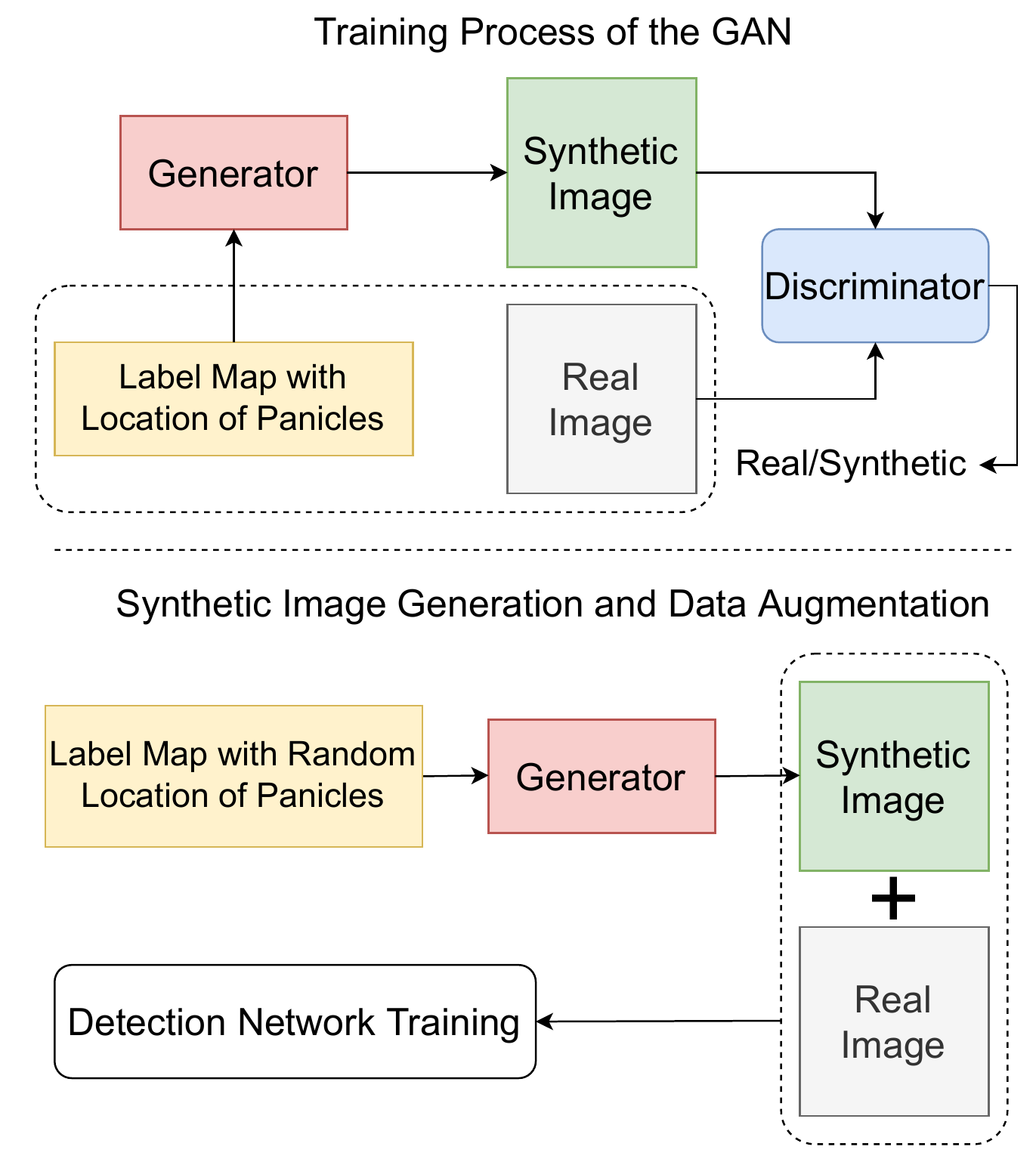}
    \caption{Block diagram of our proposed approach.}
    \label{fig:flow}
\end{figure}

Our method consists of two parts: training a conditional GAN with a small amount of ground truth UAV RGB images (label maps) of Sorghum and then generating synthetic images with random locations of the panicles which we use for data augmentation.
\Cref{fig:flow} illustrates our proposed approach.
In the top part of \Cref{fig:flow}, label maps and images are used for training the GAN.
The label map is a grayscale mask acquired by converting existing ground truth bounding box labels.
In the bottom part of \Cref{fig:flow}, we use random locations of the panicles (random label map) as an input to the GAN to generate synthetic UAV RGB images.
The synthetic images are then added to the real ground truth image dataset to form the training data we will use to train the panicle detection network.
The requirements for the GAN are that it must be capable of generating high-resolution images constrained by the inputs.
We choose two image-to-image translation GANs because the real ground image bounding box labels can be converted directly into label masks.
The two GANS we use are pix2pixHD\cite{pix2pixhd} and SPADE\cite{spade} because of their reported performance on public datasets such as ADE20K\cite{ade20k} and Cityscapes\cite{cityscape}.

\subsection{pix2pixHD}
\begin{figure}[t]
    \centering
    \includegraphics[width=0.47\textwidth]{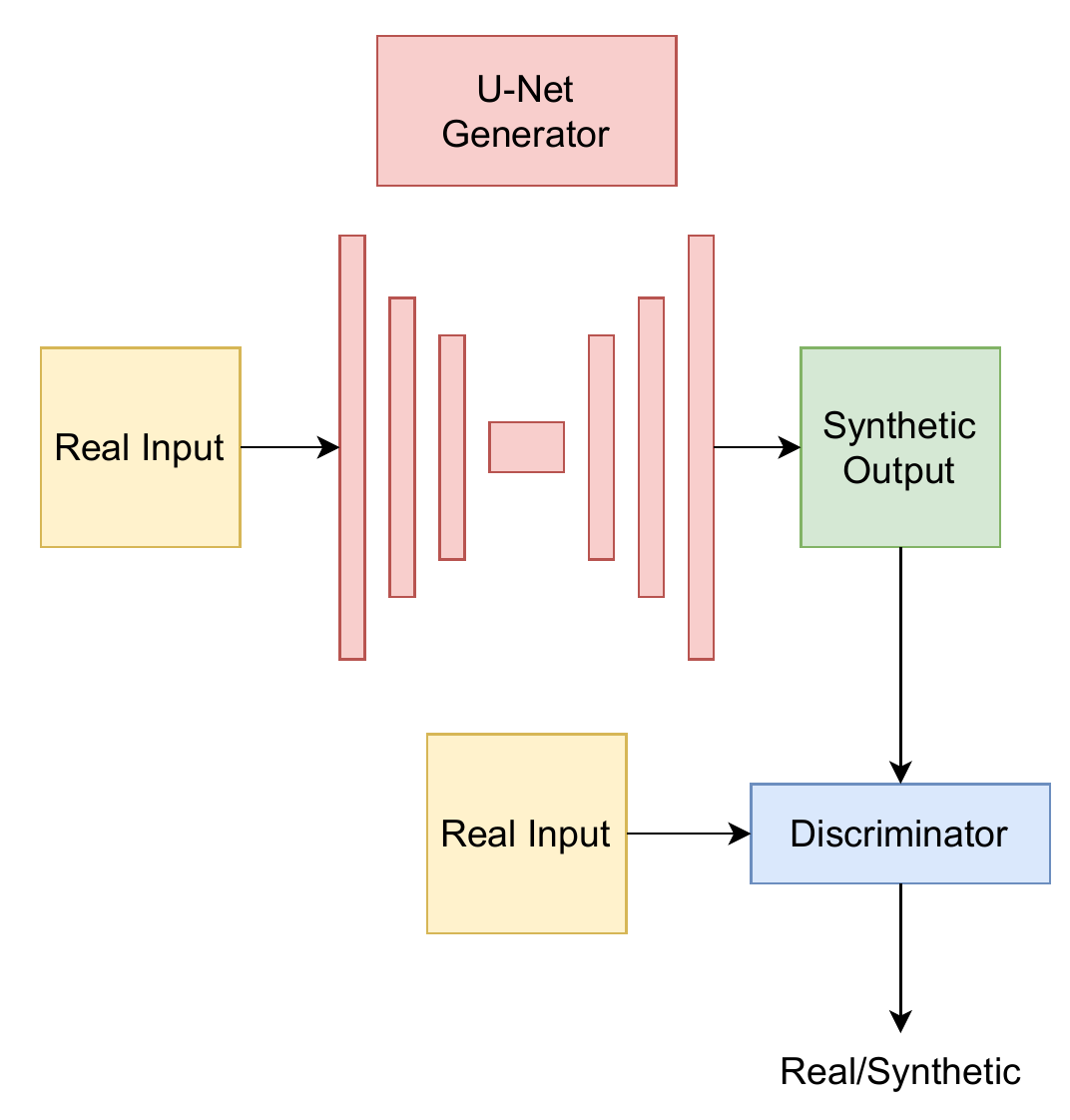}
    \caption{The pix2pix structure.}
    \label{fig:pix2pix}
\end{figure}

The pix2pixHD\cite{pix2pixhd} architecture is an image-to-image translation GAN that can generate high-resolution images from label map inputs.
The concept of pix2pixHD comes from  pix2pix \cite{pix2pix}, a conditional GAN for image-to-image translation.
The original pix2pix has the structure shown in \Cref{fig:pix2pix}.
pix2pix comprises of a generator $G$ and a discriminator $D$.
It has the following objective function:
\begin{equation}
    \mathcal{L}_{\text{GAN}}(G, D) \equiv
    \mathbb{E}_{(\mathbf{s}, \mathbf{x})}[\log D(\mathbf{s}, \mathbf{x})]+\mathbb{E}_{\mathbf{s}}[\log (1-D(\mathbf{s}, G(\mathbf{s}))]
\end{equation}
where $(s_i, x_i)$ is the $i$-th sample pair of a label and its corresponding nature image. We denote $\mathbb{E}_{(\mathbf{s}, \mathbf{x})} \triangleq \mathbb{E}_{(\mathbf{s}, \mathbf{x})} \sim p_{\text {data }}(\mathbf{s}, \mathbf{x})$.
The generator and discriminator of pix2pix are derived from the DCGAN\cite{dcgan} which uses a CNN as its main structure.
The generator of pix2pix uses skip connections from U-Net\cite{unet} to share information between input and output.
The discriminator of pix2pix adapts the PatchGAN\cite{pix2pix} structure.
The PatchGAN is a patch-based CNN that requires fewer parameters to train compared to fully connected structures of other discriminators.

\begin{figure}[t]
    \centering
    \includegraphics[width=0.45\textwidth]{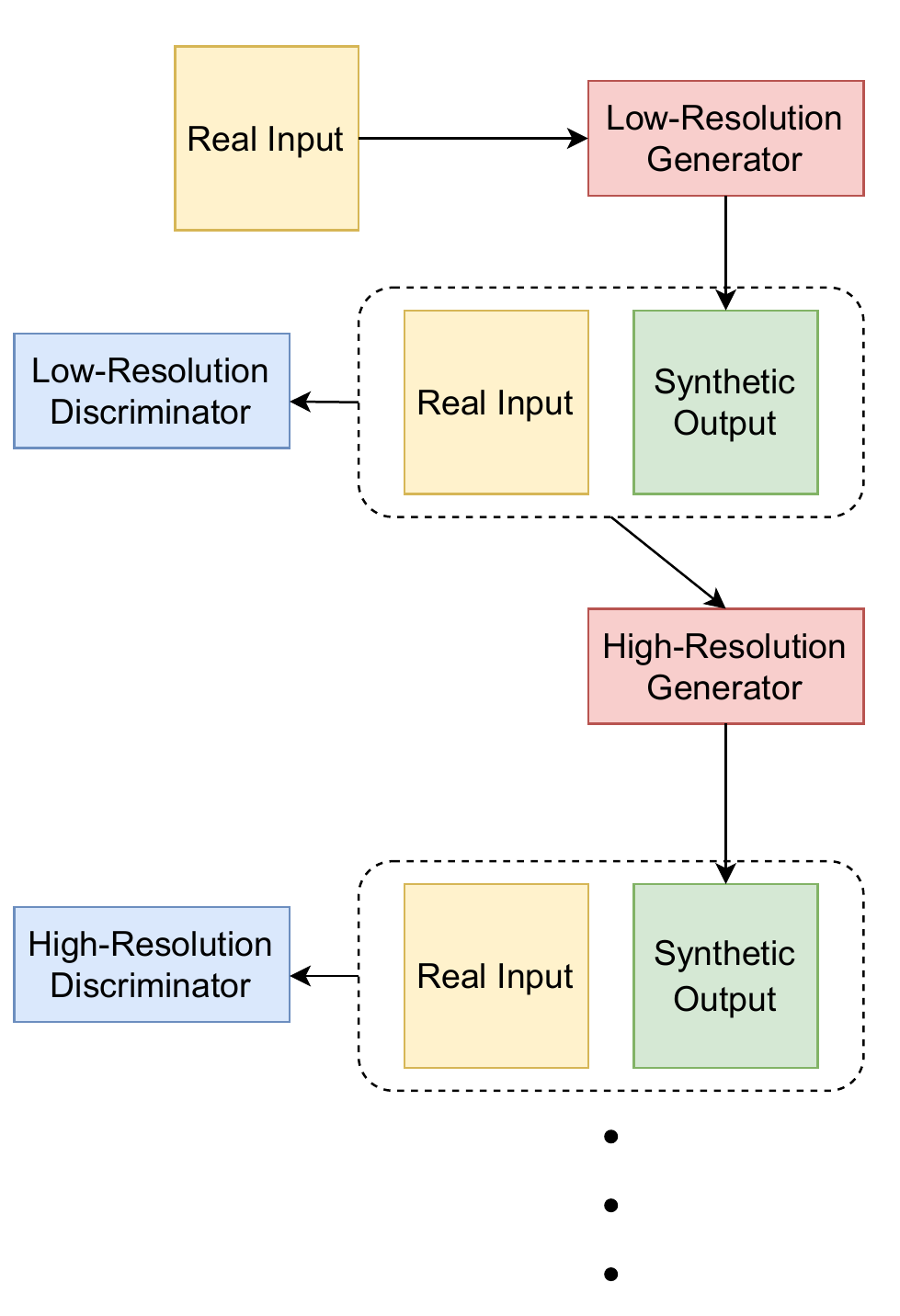}
    \caption{The pix2pixHD structure.}
    \label{fig:pix2pixhd}
\end{figure}

pix2pix can only generate images with a maximum resolution of $256 \times 256$ pixels. 
If we train pix2pix with images larger than $256 \times 256$ pixels, the training process becomes unstable and the image quality deteriorates.
pix2pixHD\cite{pix2pixhd} addresses the limitation of pix2pix by introducing multi-scale generators and discriminators.
The structure of pix2pixHD is shown in \Cref{fig:pix2pixhd}.
It has the following objective function:
\begin{equation}
    \min_{G}\Bigg(
    \Big(
    \max_{D_{1}, D_{2}, D_{3}} \sum_{\mathclap{k=1,2,3}} \mathcal{L}_{\text{GAN}}(G, D_{k})
    \Big)
    +\mathclap{\lambda} \sum_{\mathclap{k=1,2,3}} \mathcal{L}_{\text{FM}}(G, D_{k})\Bigg)
\end{equation}
where $\lambda$ controls the importance of a GAN loss $\mathcal{L}_{\text{GAN}}(G, D_{k})$ and a feature matching loss $\mathcal{L}_{\mathrm{FM}}(G, D_{k})$.
The $k$-th feature matching loss $\mathcal{L}_{\mathrm{FM}}(G, D_{k})$ for the discriminator is incorporated in pix2pixHD to stabilize training:
\begin{equation}
    \mathcal{L}_{\mathrm{FM}}(G, D_{k})\equiv
    \mathbb{E}_{(\mathbf{s}, \mathbf{x})}
    \sum_{i=1}^{T} \frac{1}{N_{i}}||D_{k}^{(i)}(\mathbf{s}, \mathbf{x})-D_{k}^{(i)}(\mathbf{s}, G(\mathbf{s}))||_{1}
\end{equation}
where $D_{k}^{(i)}$ is the feature extractor of discriminator $D_k$ for $i$-th layer, $N_i$ is denotes the number of elements in the $i$-th layer and $T$ denotes the total number of layers.

The multiscale generators and discriminators form pyramid structures for synthesizing images with different resolutions.
The generator structures of pix2pixHD consist of a global generator $G_{1}$ and a local enhancer network $G_{2}$.
During training, $G_{1}$ is first trained on low-resolution images.
$G_{2}$ is appended to the last layer of $G_{1}$ and trained jointly on high-resolution images.
The discriminator structures of pix2pixHD are paired with multiscale generators for low-resolution and high-resolution images.
The input of its discriminator is real images and synthetic images that are downsampled at multiple scales that match the pyramid structures.
The high-resolution discriminator can encourage generating images with finer details.
The low-resolution discriminator guides the generator to generate globally consistent images.
pix2pixHD also uses an optional instance map as an auxiliary input. 
Each individual object inside the instance map has a unique ID.
The instance map provides the object  boundaries of the same class.
However, for our application creating panicle segmentation masks for small objects is very time-consuming, which conflicts with our goal of saving time in ground truthing.
We decide not to use instance maps for generating Sorghum images.

\begin{figure}[t]
    \centering
    \includegraphics[width=0.35\textwidth]{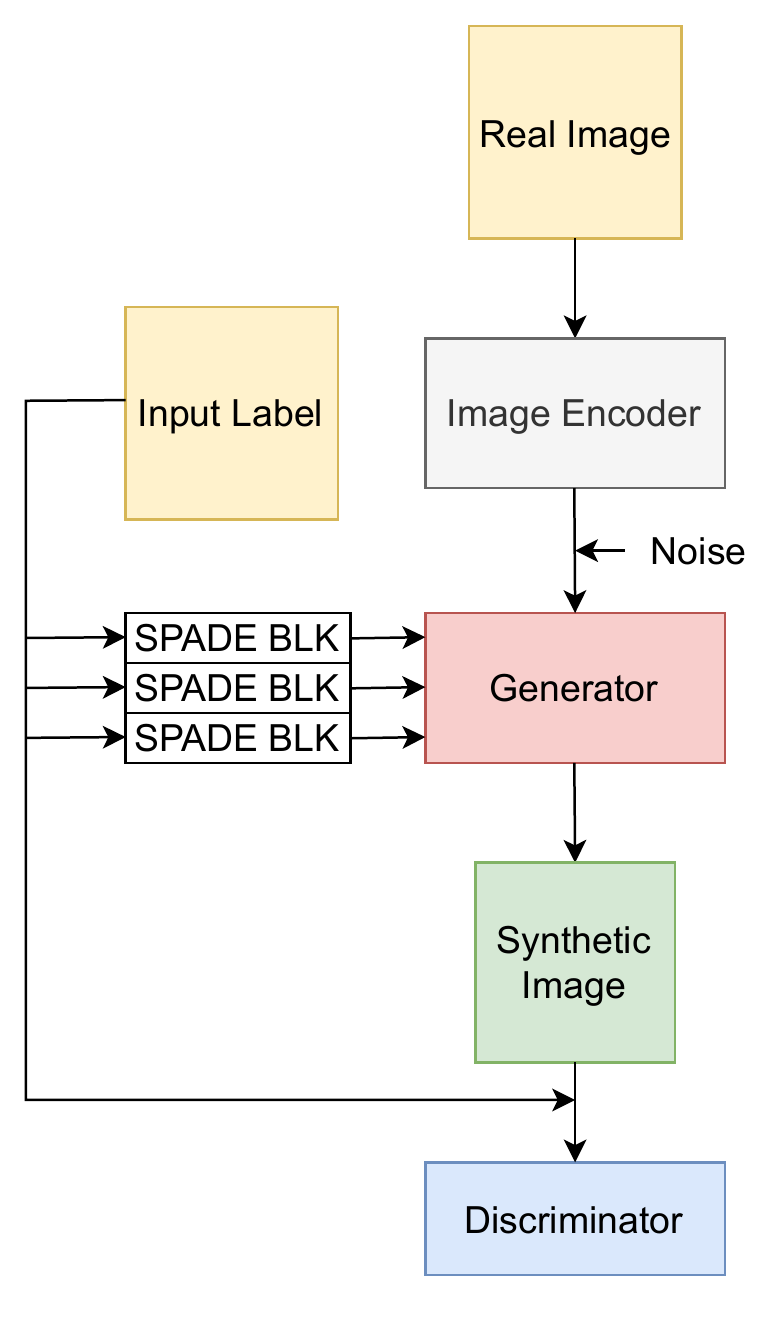}
    \caption{The SPADE structure.}
    \label{fig:spade}
\end{figure}

\subsection{SPADE}
The regular normalization layers from the generator of pix2pixHD tend to lose semantic information.
The diminished semantic information affects the quality of the synthetic images.
Spatially-adaptive denormalization (SPADE)\cite{spade}  preserves the semantic information compared to regular normalization by incorporating residual blocks with SPADE\cite{spade} layers into the up-sampling part of the generators as shown in \Cref{fig:spade}.
The SPADE layer consists of a convolutional layer and two modulation layers.
The convolutional layer during training uses  ground truth label mask as input.
Two modulation layers take the output of the convolutional layer and output two modulation parameters $\gamma$ and $\beta$.
The $\gamma$ and $\beta$ are multiplied and added to the normalized activation as output.

The generator of SPADE discards the encoder structures from pix2pixHD since the label maps are directly fed into the residual blocks within the generators.
A new image encoder is used to encode the real image into random vectors before sending it into the up-sampling generator.
The discriminator of SPADE follows the multi-scale discriminators of pix2pixHD with SPADE layers as normalization.
The discriminator takes the concatenation of the input ground truth mask and the synthetic images and classifies whether the generated image is synthetic.

\begin{figure}[t]
    \centering
    \includegraphics[width=0.47\textwidth]{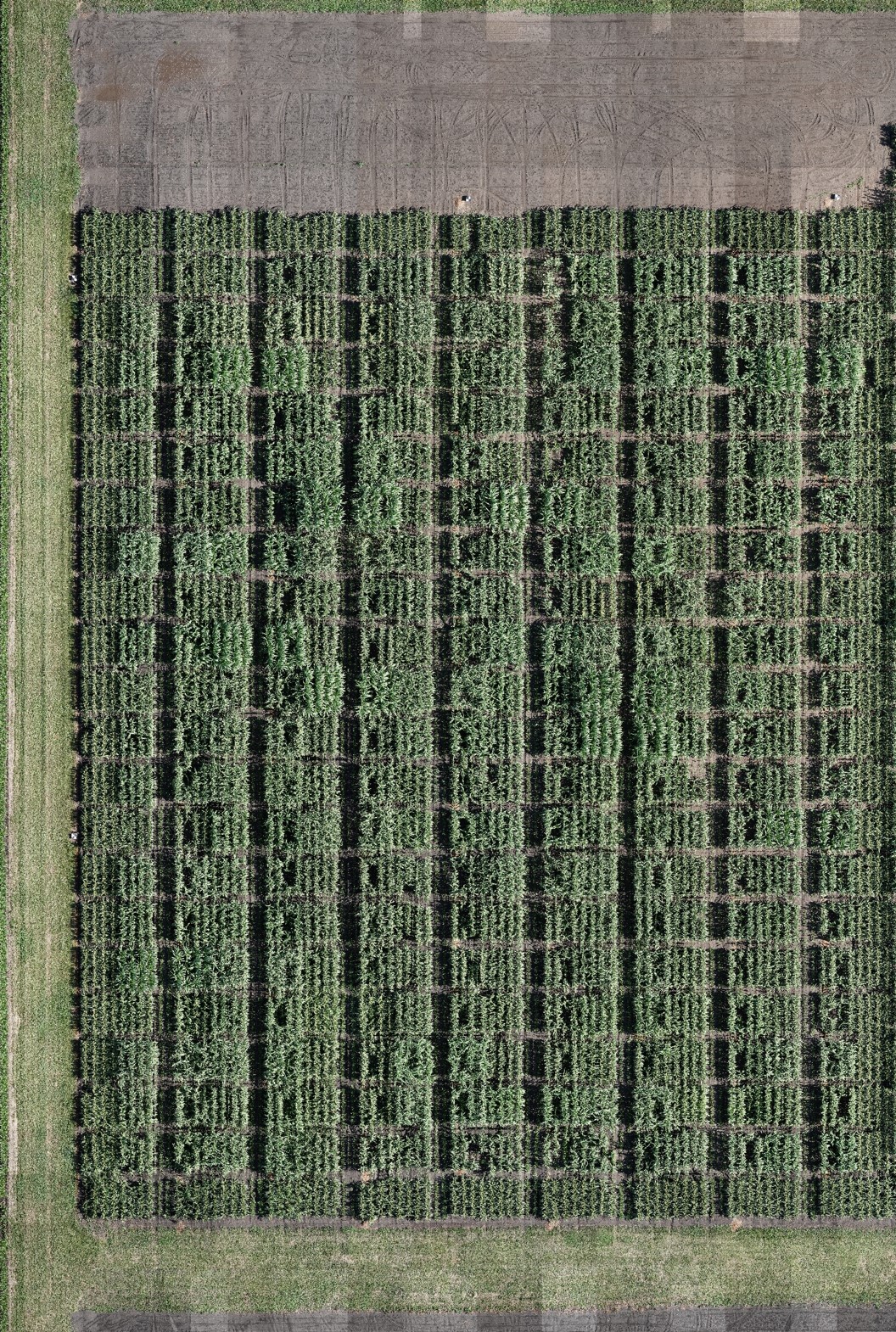}
    \caption{An orthomosaic of a Sorghum field. The dataset used for the experiments is cropped from a set of orthomosaics.}
    \label{fig:ortho}
\end{figure}

\begin{figure*}[t]
    \centering
    \begin{subfigure}[t]{0.24\textwidth}
        \includegraphics[width=\textwidth]
        {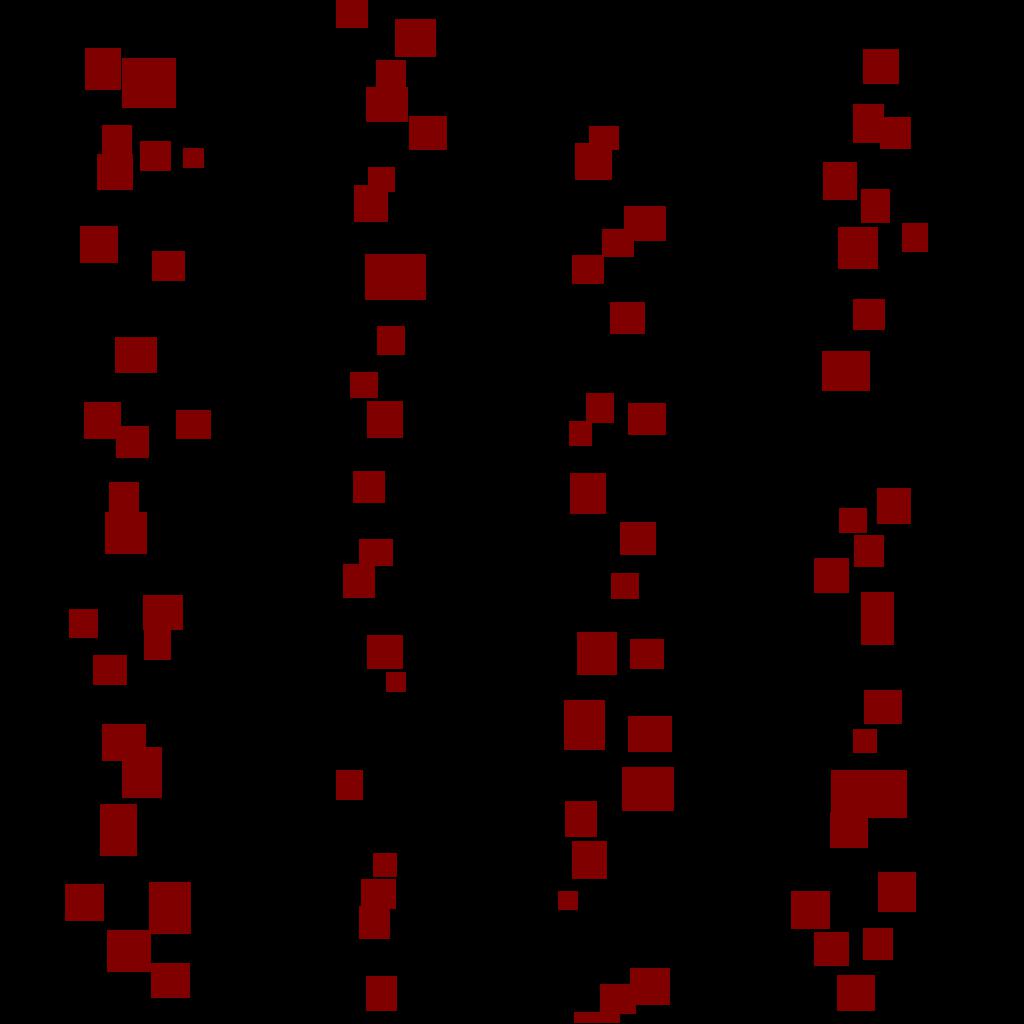}
        \includegraphics[width=\textwidth]
        {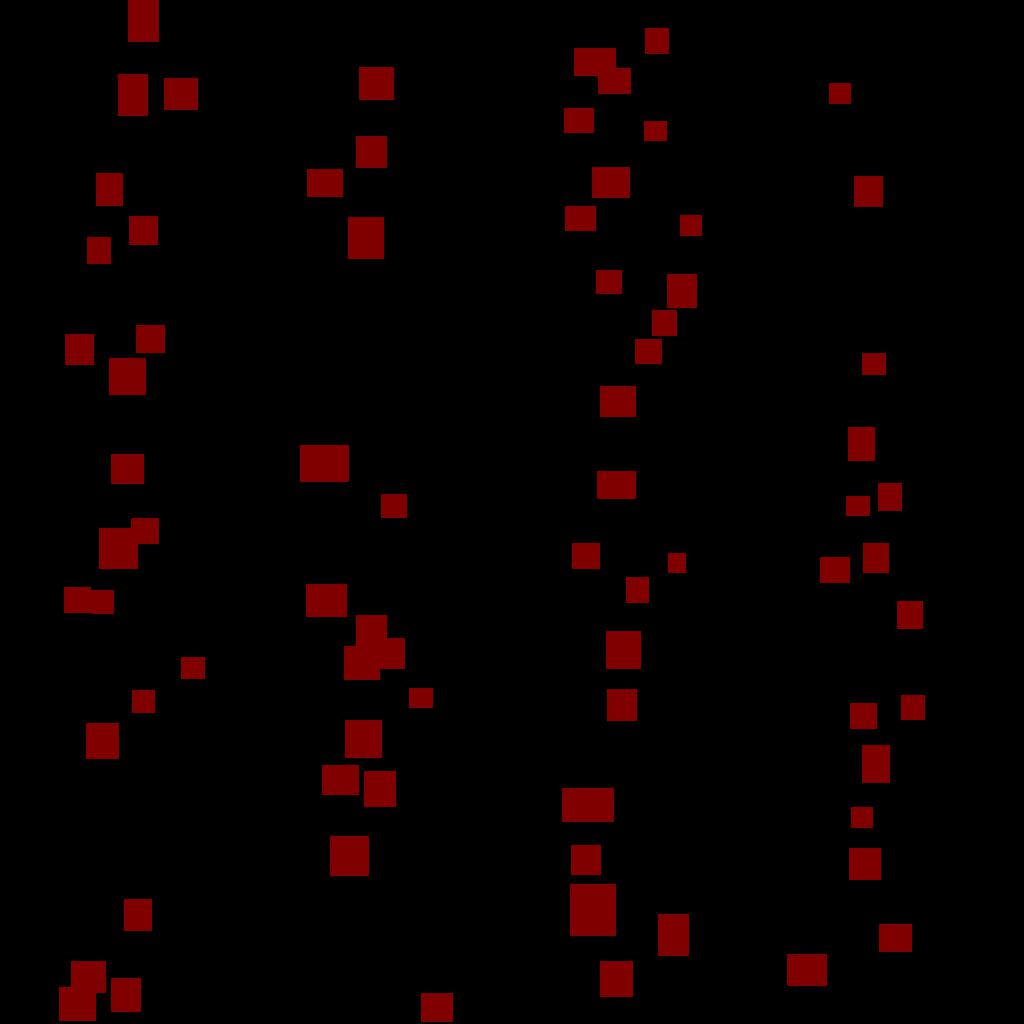}
        \includegraphics[width=\textwidth]
        {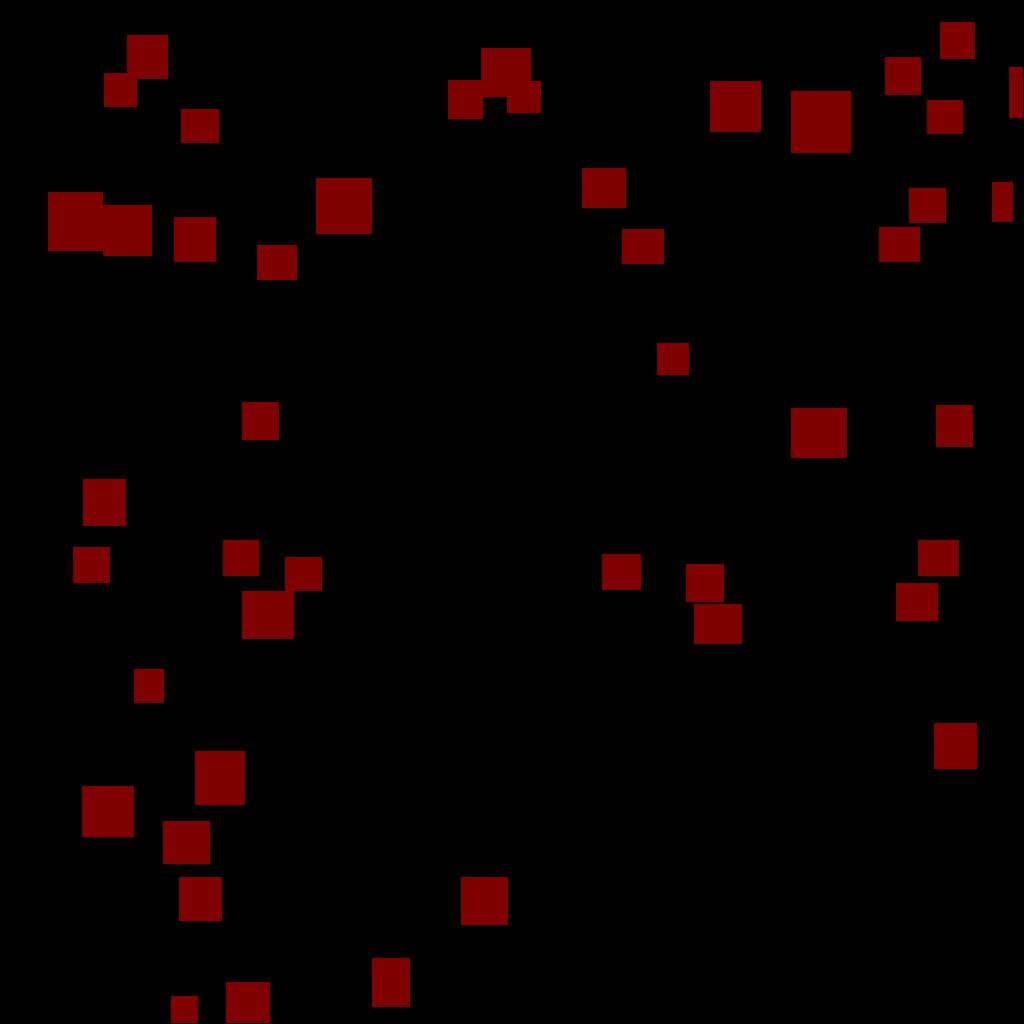}
        \caption{Ground Truth Label map}
    \end{subfigure}\hfill
    \begin{subfigure}[t]{0.24\textwidth}
        \includegraphics[width=\textwidth]  
        {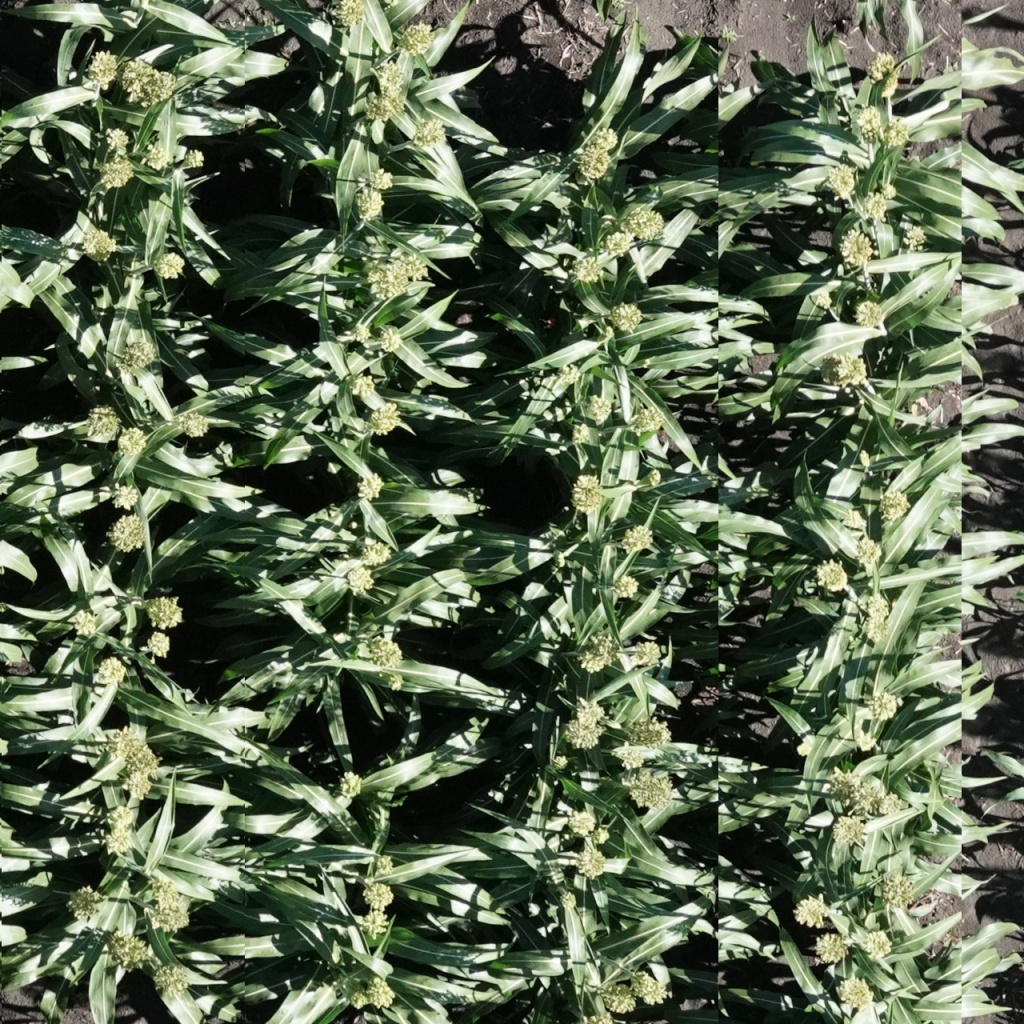}
        \includegraphics[width=\textwidth]
        {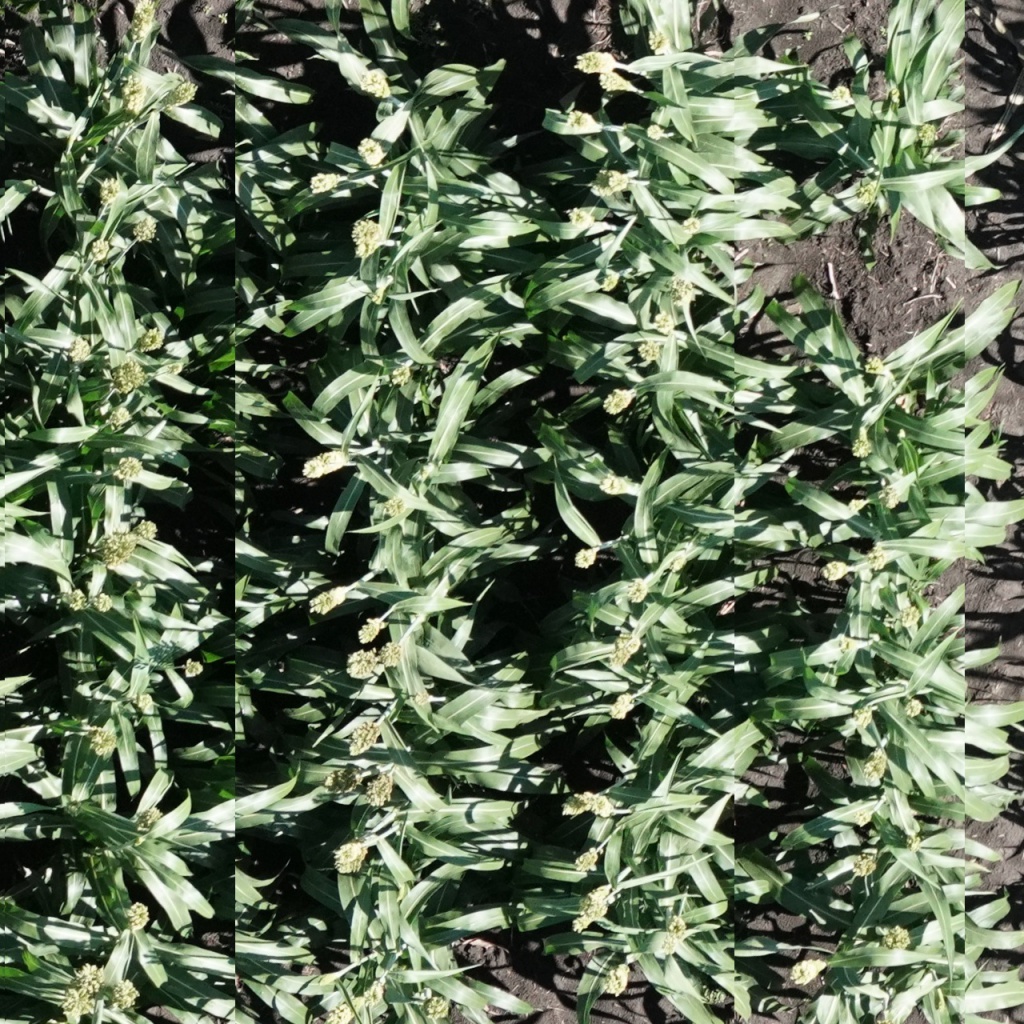}
        \includegraphics[width=\textwidth]
        {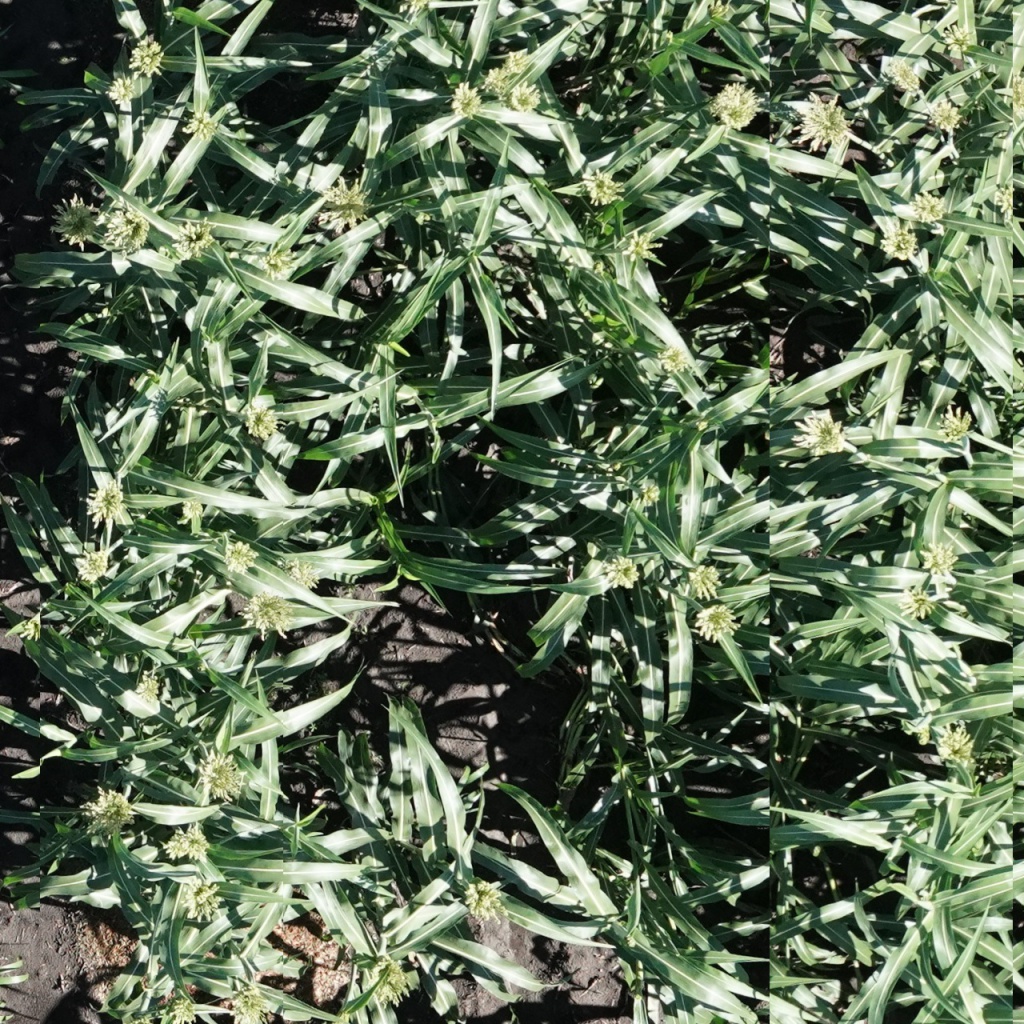}
        \caption{Real images}
    \end{subfigure}\hfill
    \begin{subfigure}[t]{0.24\textwidth}
        \includegraphics[width=\textwidth]  
        {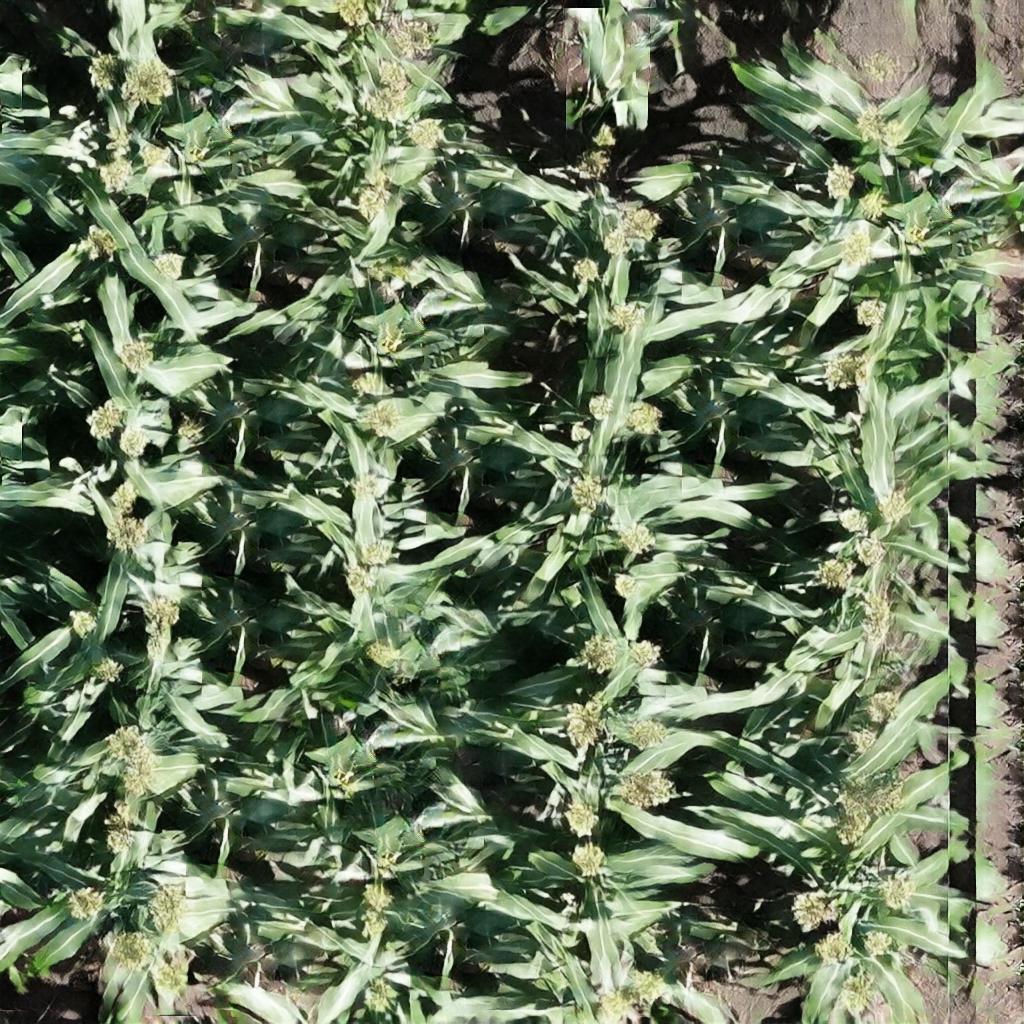}
        \includegraphics[width=\textwidth]
        {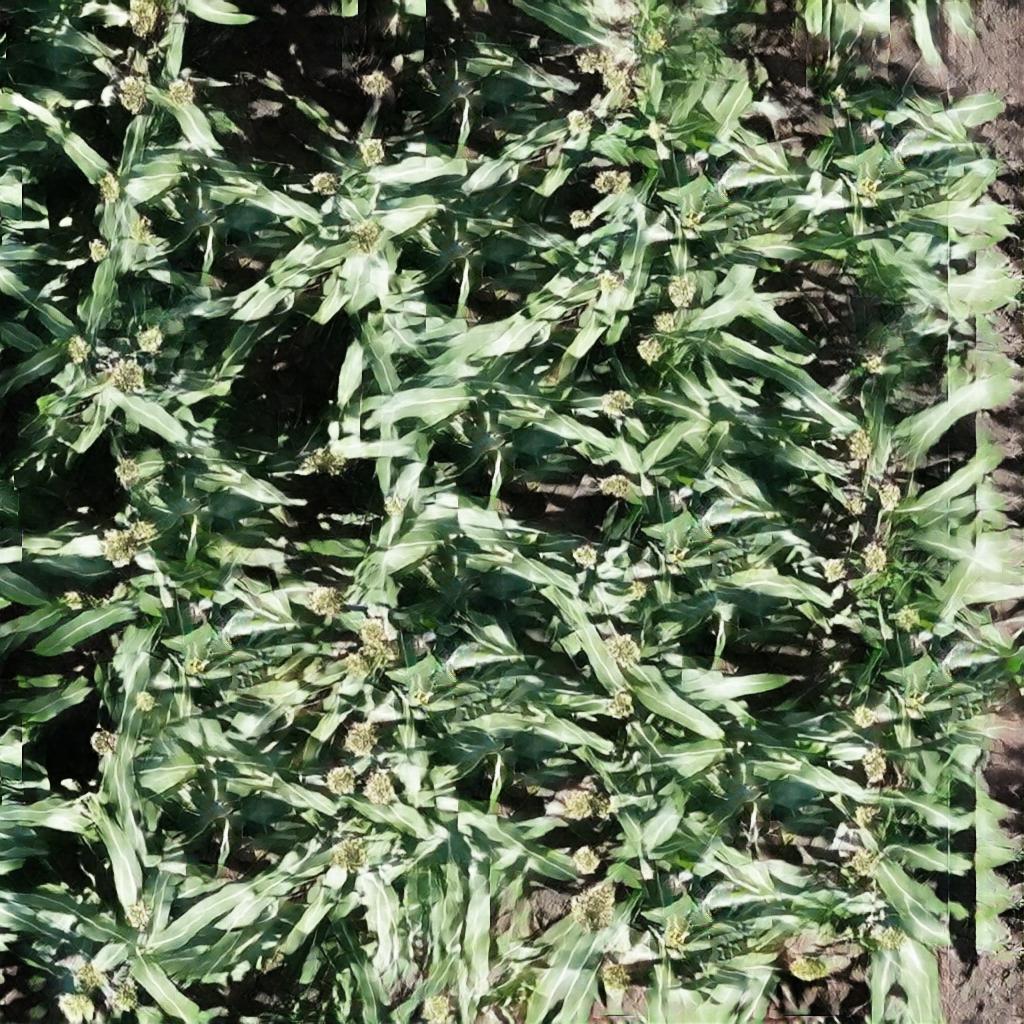}
        \includegraphics[width=\textwidth]
        {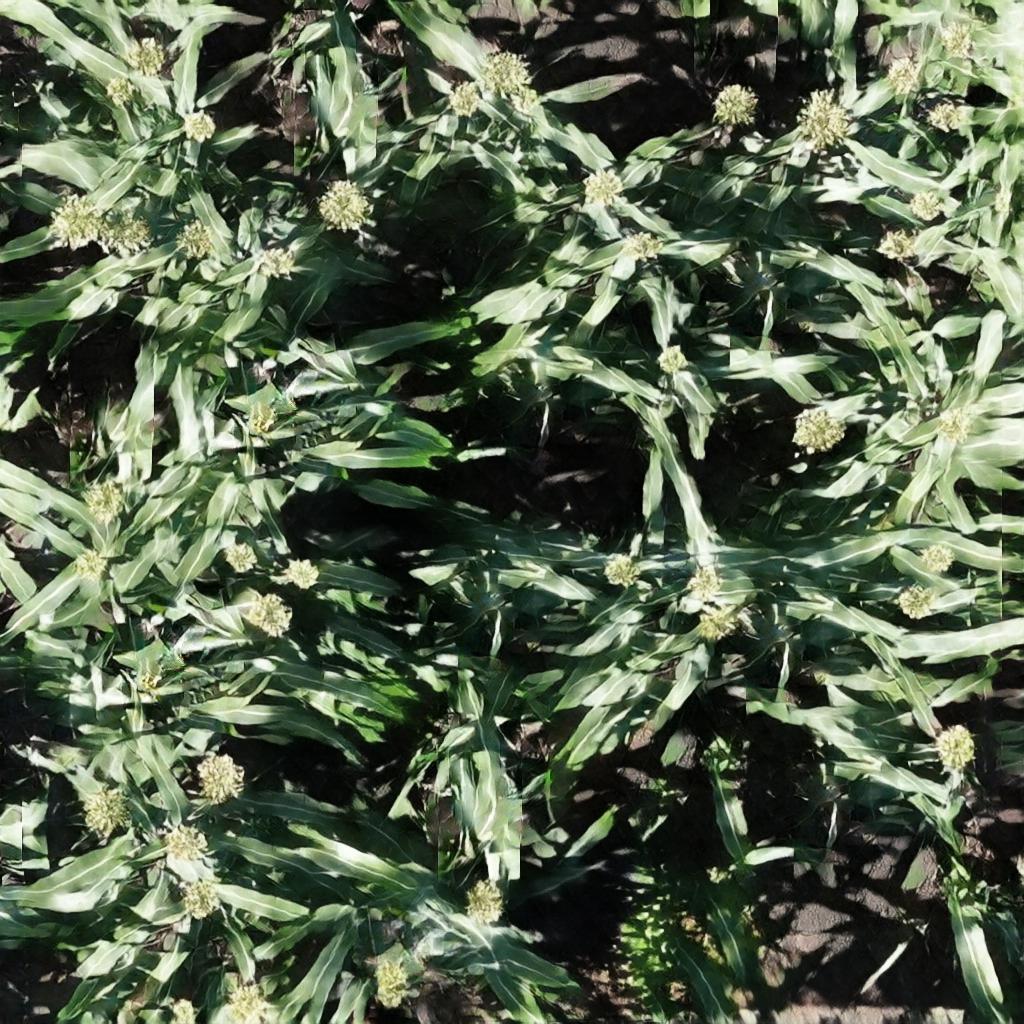}
        \caption{pix2pixHD results}
    \end{subfigure}\hfill
    \begin{subfigure}[t]{0.24\textwidth}
        \includegraphics[width=\textwidth]  
        {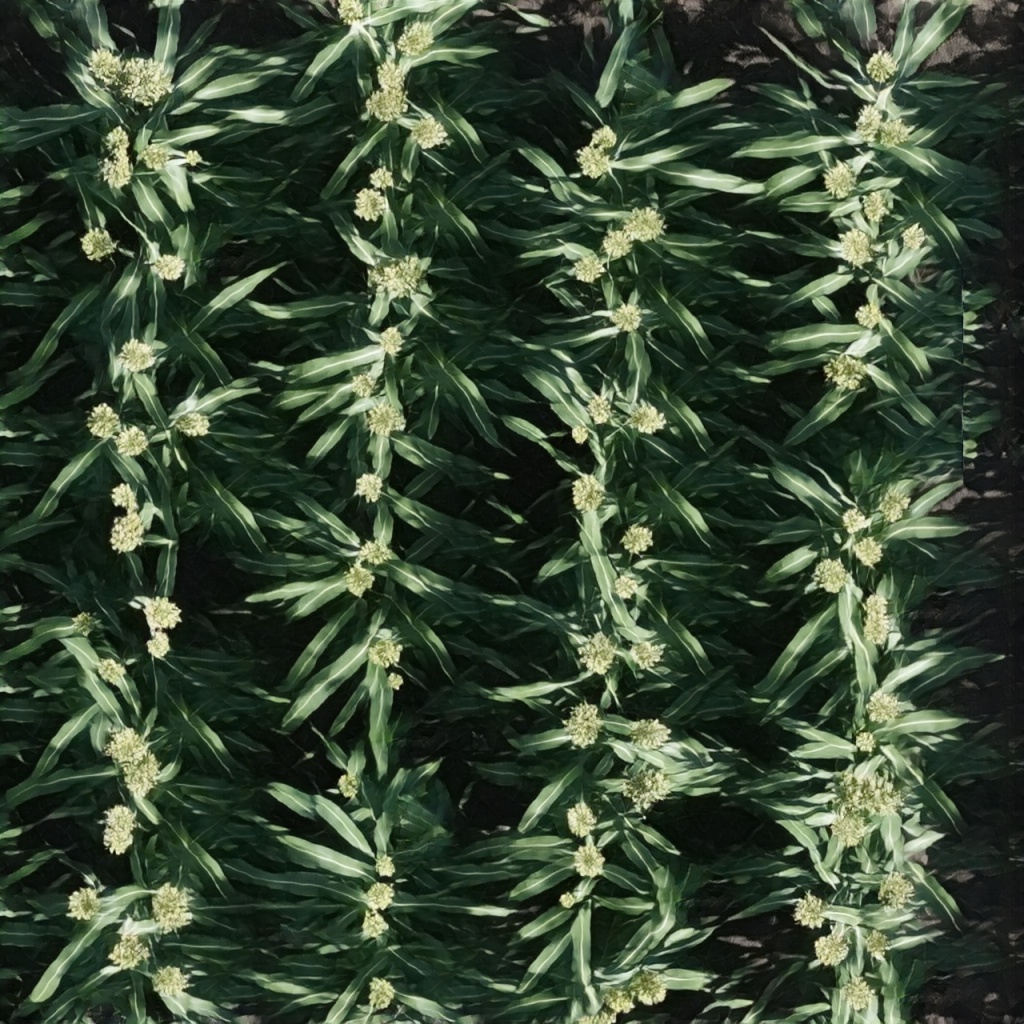}
        \includegraphics[width=\textwidth]
        {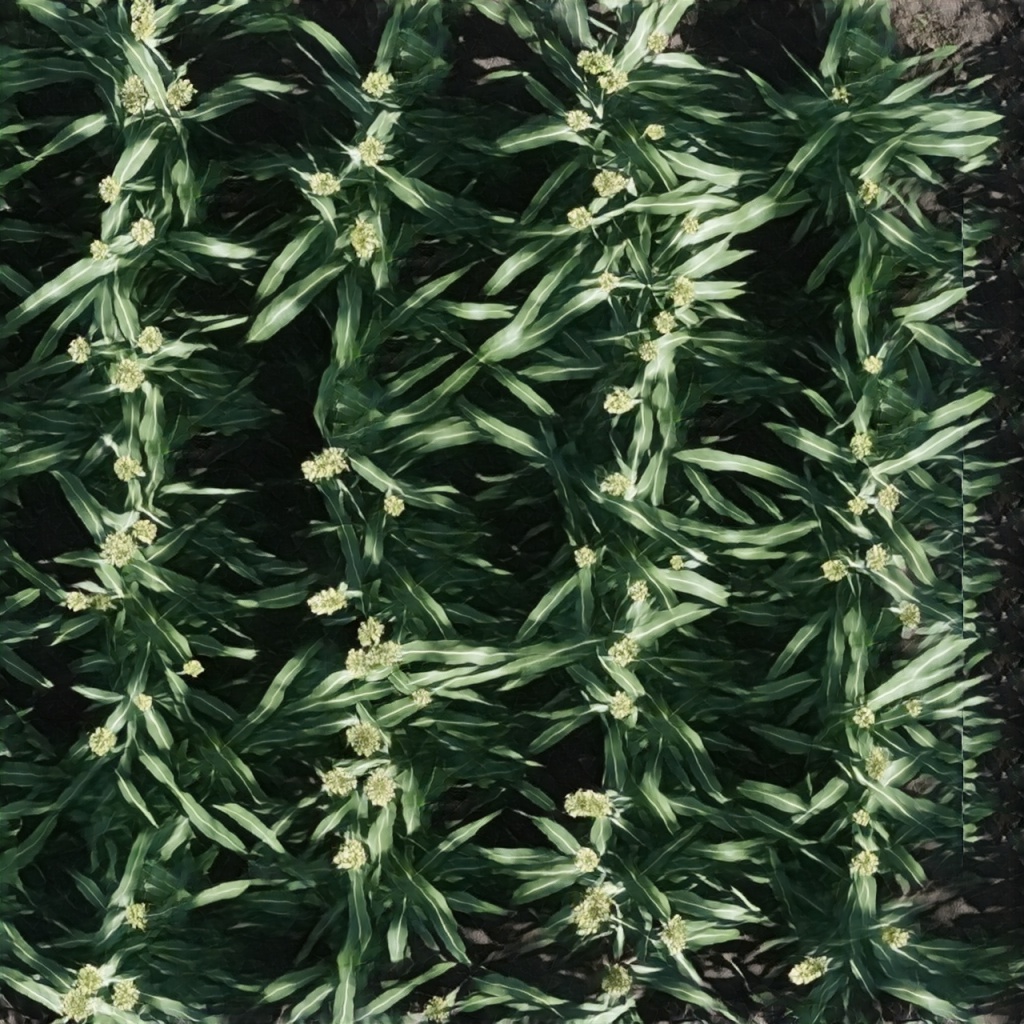}
        \includegraphics[width=\textwidth]
        {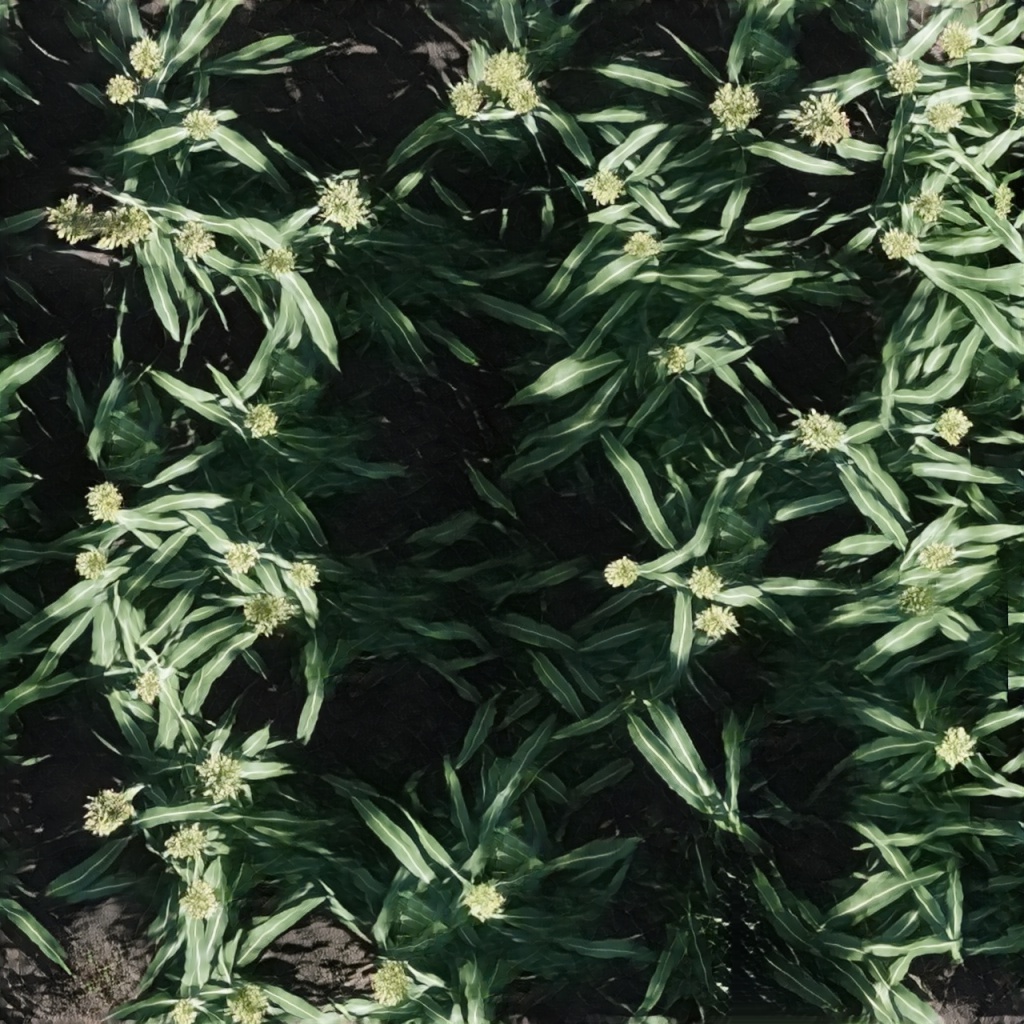}
        \caption{SPADE results}
    \end{subfigure}
    \caption{Real panicle images and synthetic panicle images generated using pix2pixHD and SPADE.
    }
    \label{fig:training_result}
\end{figure*}

Compared to pix2pixHD, SPADE has better performance on the public datasets such as ADE20K\cite{ade20k} and Cityscapes\cite{cityscape} due to the added SPADE blocks.
All of the public datasets have multiple classes and well-defined segmentation masks that contain a large amount of semantic information.
We only have one class in our dataset (panicles) with minimal semantic information so the added SPADE blocks for preserving semantic information might not be beneficial.
We use both methods, pix2pixHD and SPADE, and compare their performance for synthetic panicle image generation.

\begin{figure*}[t]
    \centering
    \begin{subfigure}[t]{0.30\textwidth}
        \includegraphics[width=\textwidth]
        {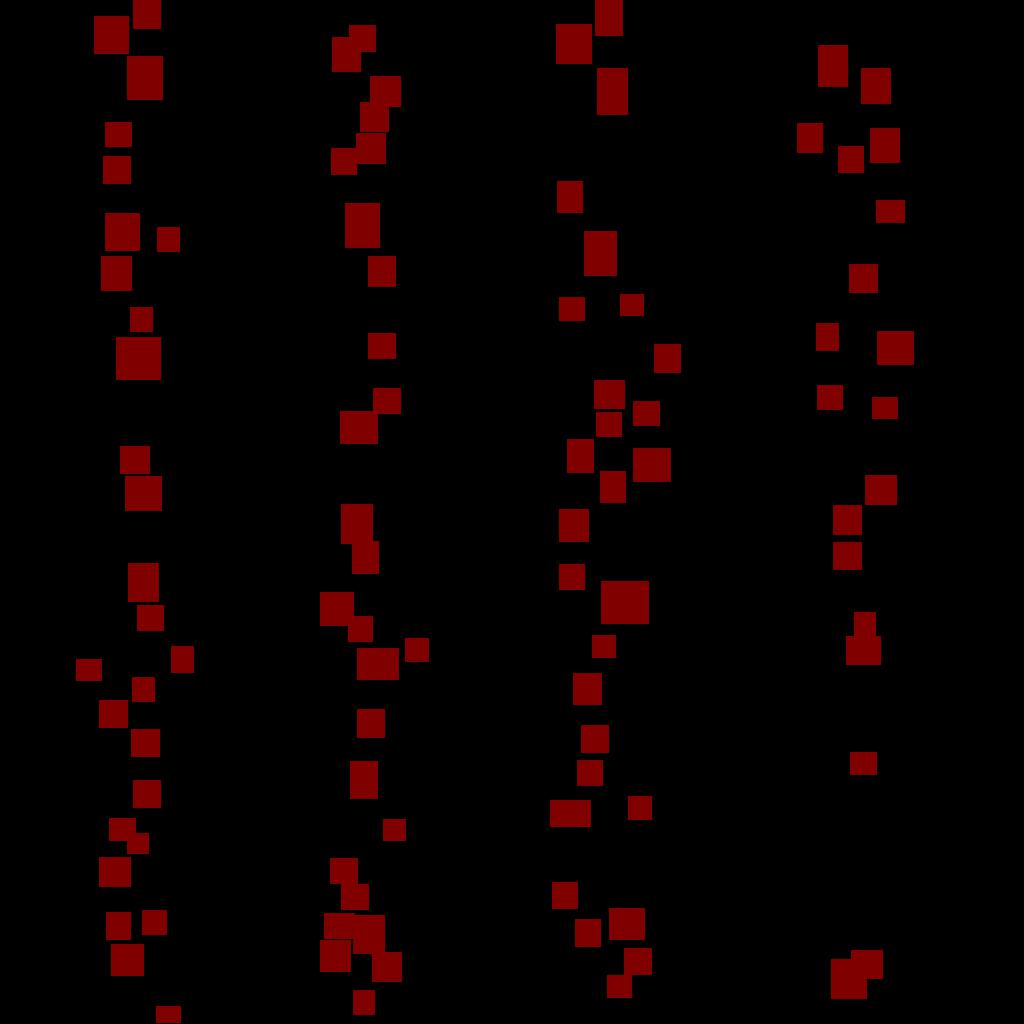}
        \includegraphics[width=\textwidth]
        {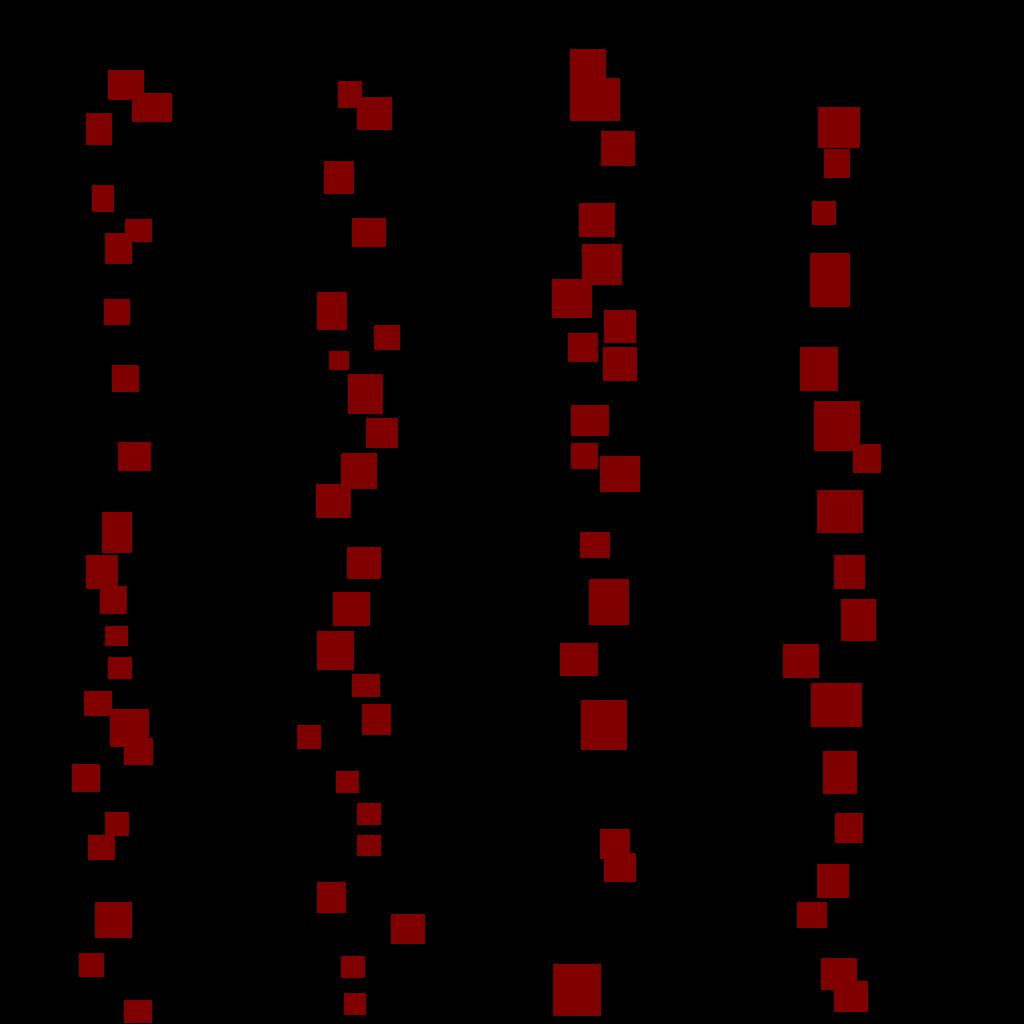}
        \includegraphics[width=\textwidth]
        {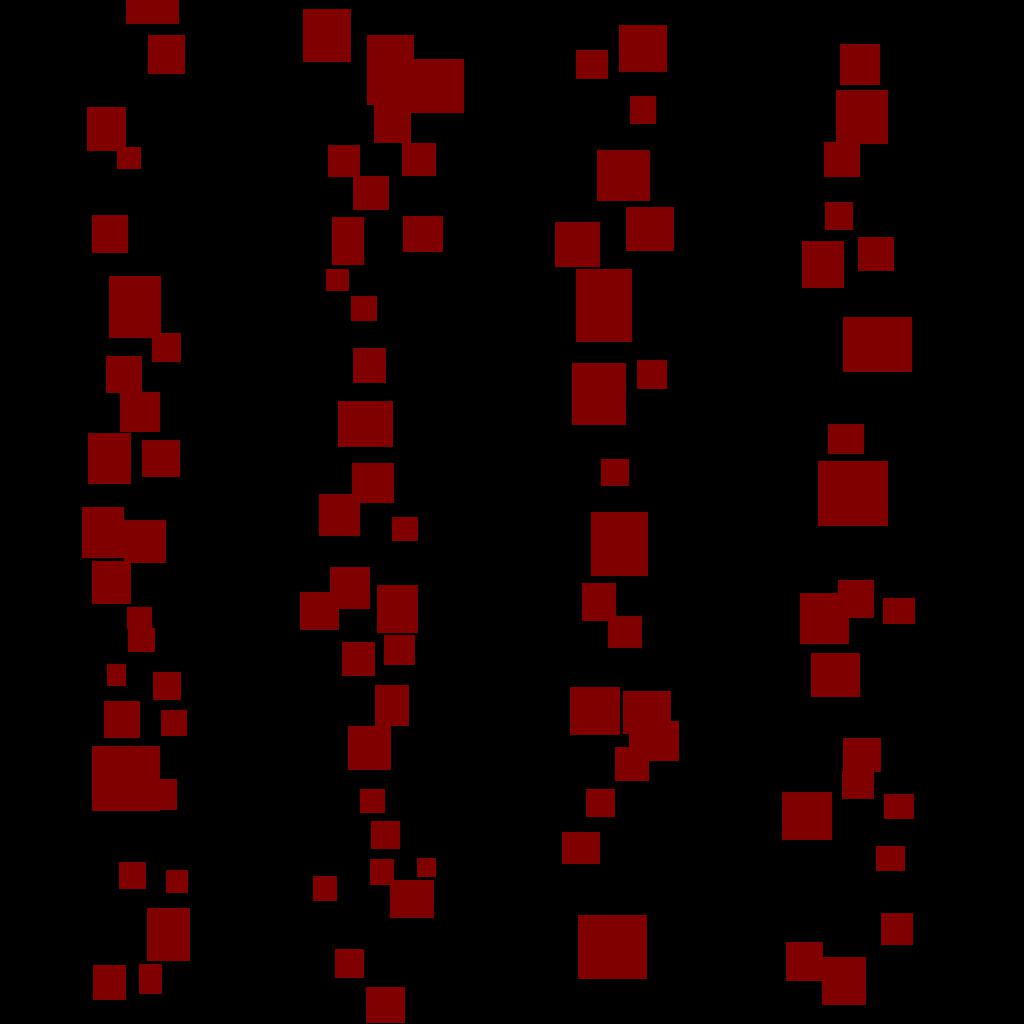}
        \caption{Random label map}
    \end{subfigure}\hfill
    \begin{subfigure}[t]{0.30\textwidth}
        \includegraphics[width=\textwidth]  
        {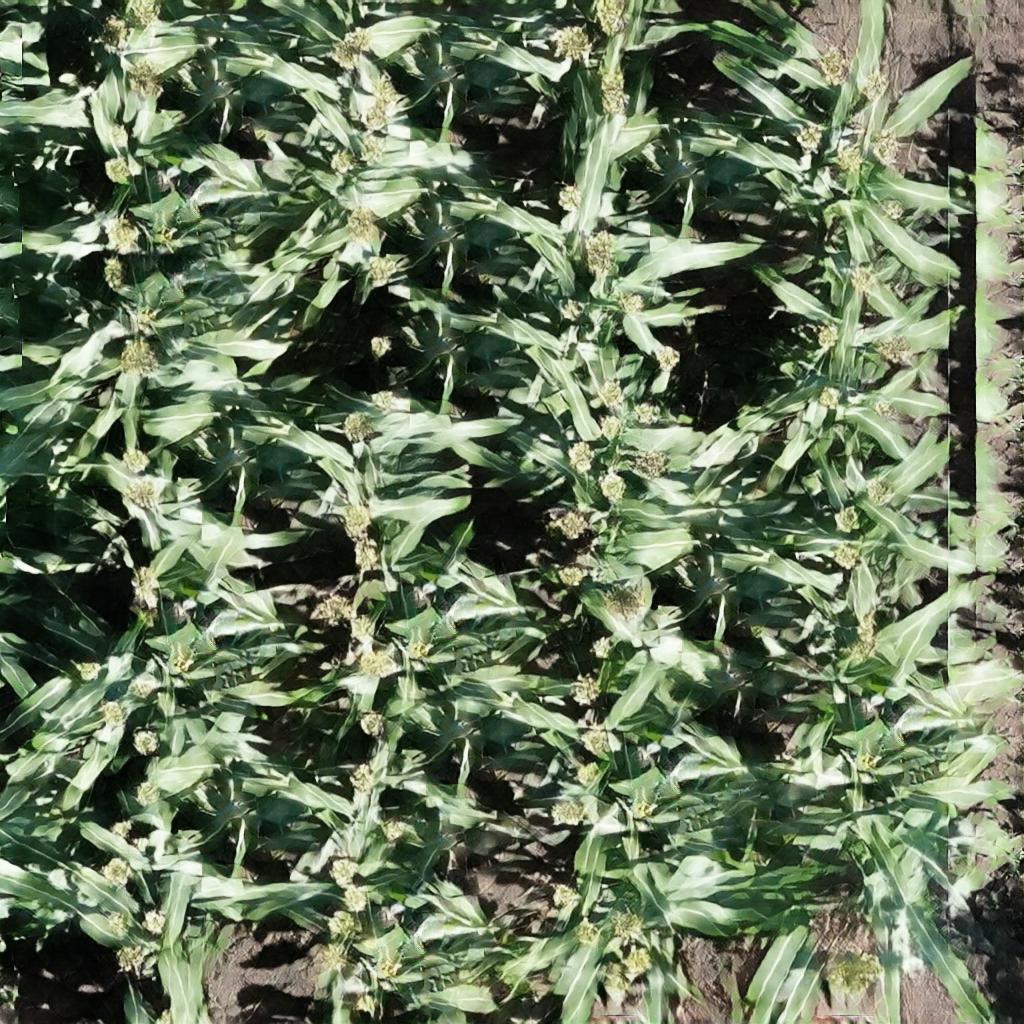}
        \includegraphics[width=\textwidth]
        {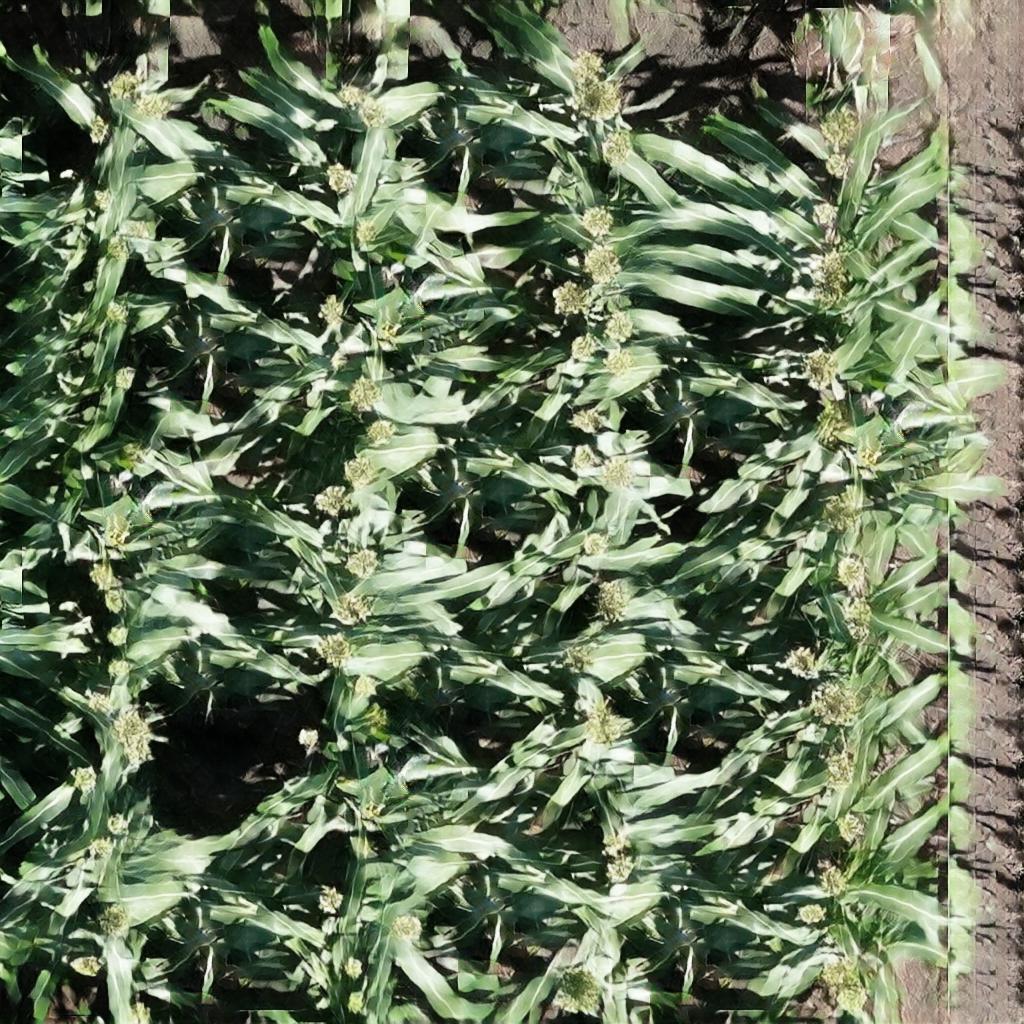}
        \includegraphics[width=\textwidth]
        {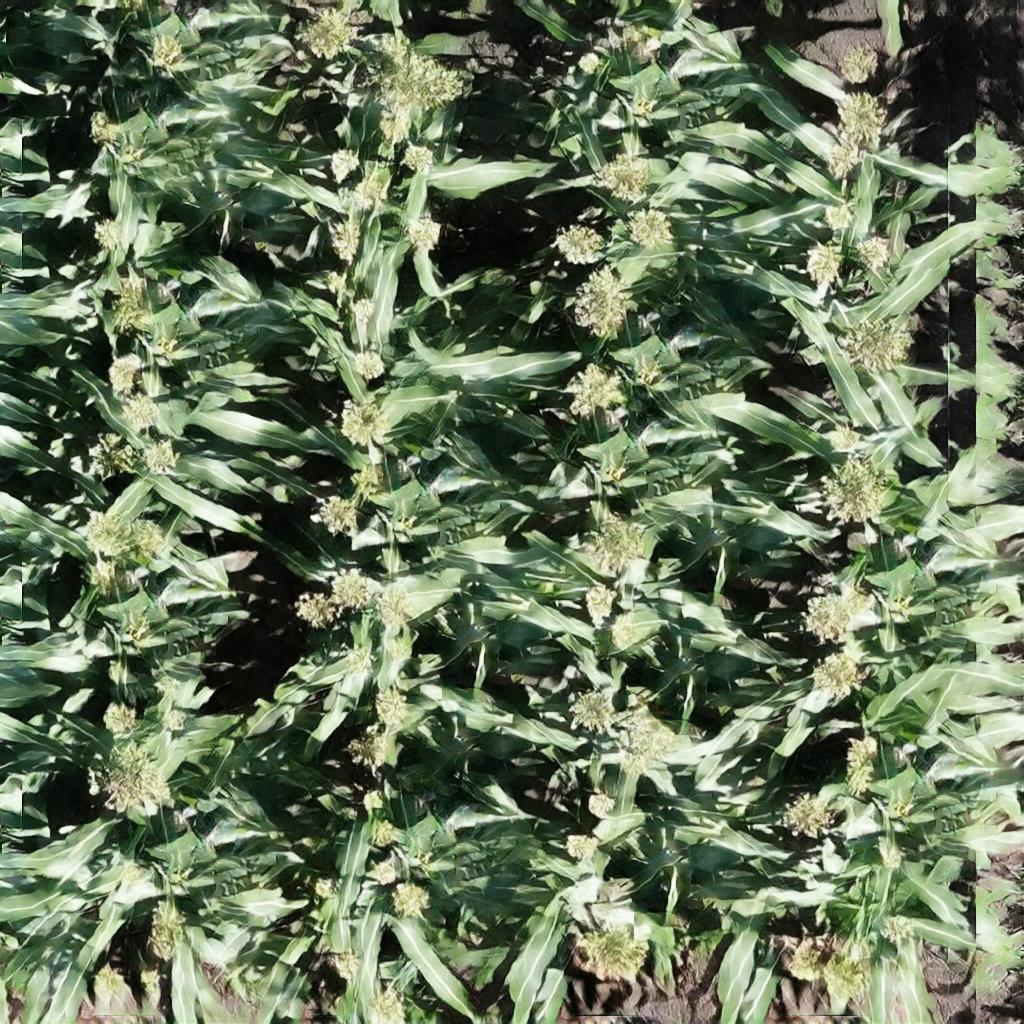}
        \caption{pix2pixHD results}
    \end{subfigure}\hfill
    \begin{subfigure}[t]{0.30\textwidth}
        \includegraphics[width=\textwidth]  
        {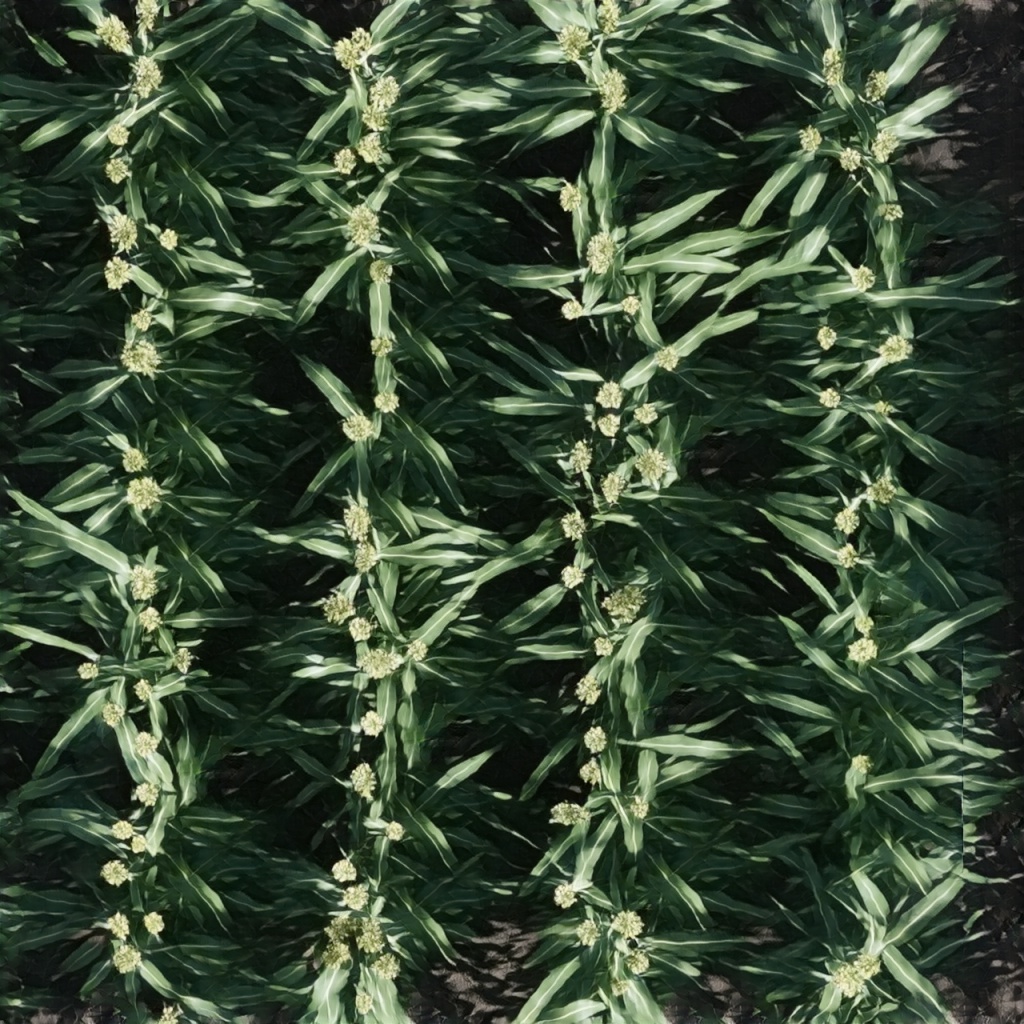}
        \includegraphics[width=\textwidth]
        {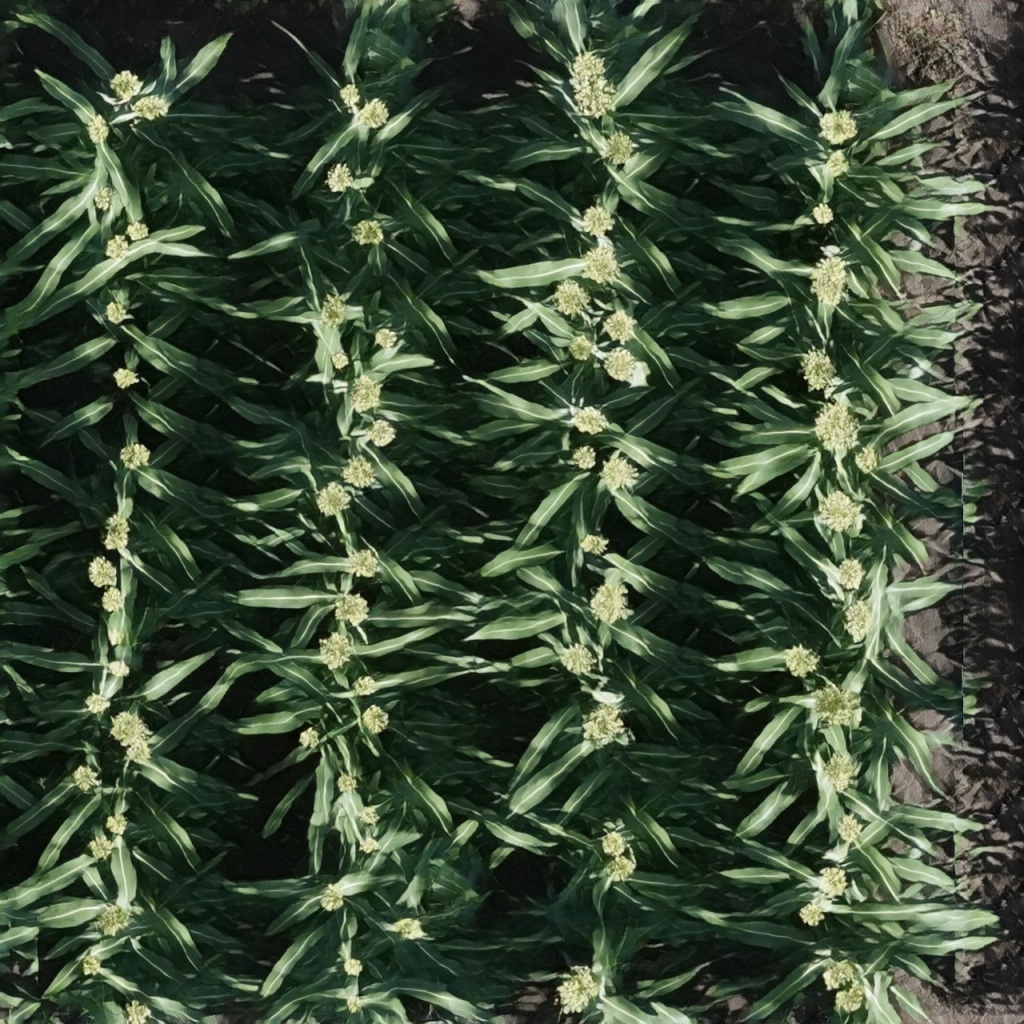}
        \includegraphics[width=\textwidth]
        {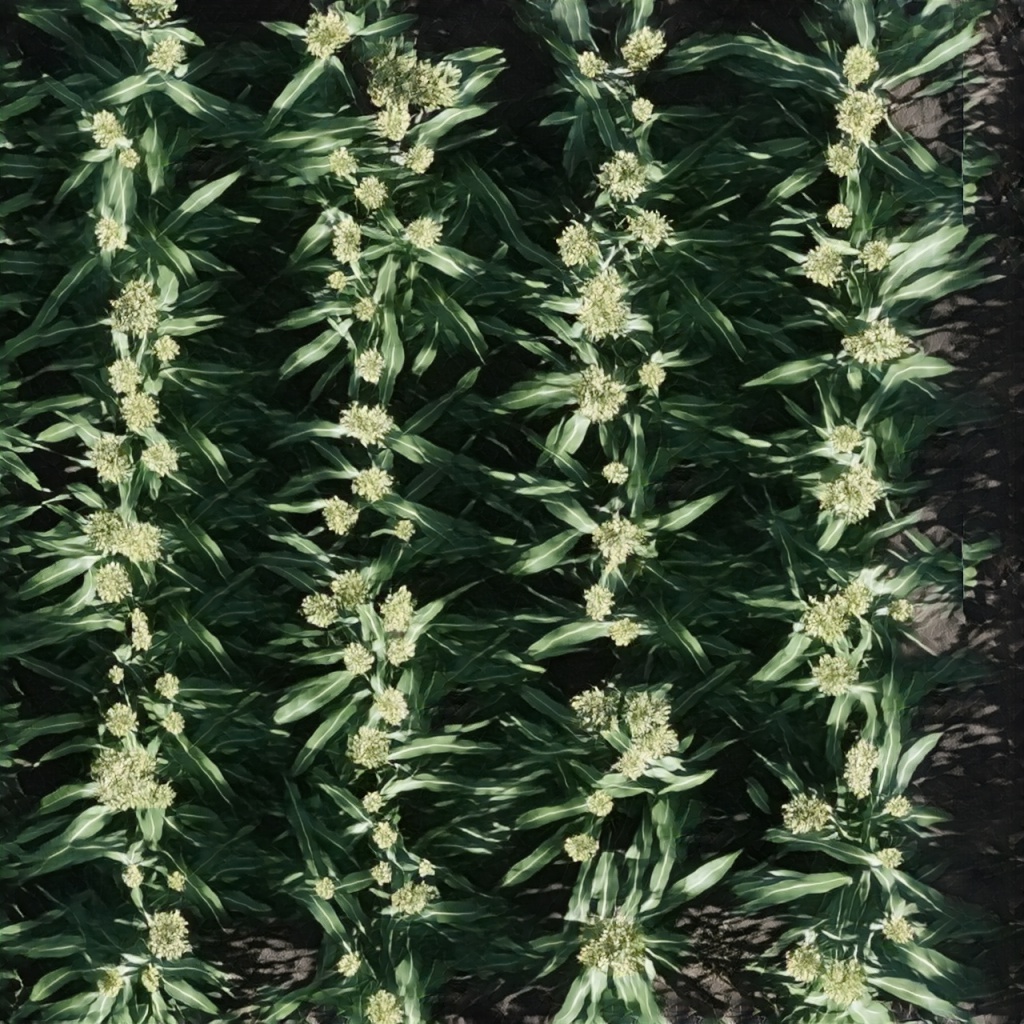}
        \caption{SPADE results}
    \end{subfigure}
    \caption{Synthetic panicle images generated using pix2pixHD and SPADE with random input labels.}
    \label{fig:panicle_result}
\end{figure*}

\section{Experiments}
\label{sec:result}
\subsection{Dataset}
The real image dataset consists of a set of cropped images from 4 orthomosaics of Sorghum field in acquired in 2020 as shown in \Cref{fig:ortho}.
The image is acquired by RGB cameras mounted on UAVs flying at an altitude of 20m.
The spatial resolution of the orthomosaics is 0.25cm/pixel.
There is a total of 4 orthomosaics from multiple dates that represent different growing stages of Sorghum.
The resolution of the cropped image is $1024 \times 1024$ pixels.
Each of the cropped images contains 4 row segments of Sorghum plants in the vertical direction.
Each panicle in the images is manually labeled with a bounding box.
The bounding box labels are converted to the ground truth label maps for training the GANs.
The label map is a single channel grayscale image that has the same resolution as the image.
The conversion has the following equation:
\begin{equation}
    p_{map}=\left\{\begin{matrix}
1,\; p_{map}\in R\\ 
0, p_{map}\notin  R
\end{matrix}\right.
\end{equation}
where $p_{map}$ is the pixel in the label map. $R$ is the region inside the bounding box.
The pixel in the label map has a value of 1 if the pixel is inside the bounding box.
The rest of the pixel has a value of 0.
In total, we have 500 manually labeled, cropped images from 4 orthomosaics as our experimental dataset.
We use 400 images for training for the GANS and 100 images for testing the panicle detection and counting system.
The number of cropped images from each orthomosaic is evenly distributed.

\subsection{Synthetic Panicle Generation}
We train pix2pixHD and SPADE with the training dataset described above.
Both methods are trained on a single NVIDIA TITAN RTX with 24GB memory and
200 epochs in native resolution ($1024 \times 1024$ pixels) without any resizing or cropping.
Adam optimization\cite{kingma_2014} is used for both pix2pixHD and SPADE with an initial learning rate of 0.0002.
The results are shown in \Cref{fig:training_result}.
The panicles generated by SPADE are more realistic than those generated by the pix2pixHD.
The panicle edges from SPADE are clearer while the panicle edges of pix2pixHD are blurred.
However, the pix2pixHD has a more natural background that resembles real Sorghum images.
The synthetic background, e.g. the highlights on leaves and the overall brightness of SPADE, is too dark.
The straight line on the right side of the synthetic images is learned from the real images created by the process of orthorectification.
Overall, the images generated by pix2pixHD are more realistic.

\subsection{Panicle Detection}
We generated 1000 synthetic images each from pix2pixHD and SPADE using random label maps (a total of 2000 GAN generated images).
\Cref{fig:panicle_result} shows some of the examples of the synthetic images.
The random label maps are constrained in location and size of the bounding boxes such that the overall distribution is identical to  real Sorghum images in our dataset.
A small overlap is allowed for each bounding box mask.
In total, we have 1400 images for training our panicle detector from each of the GANs. 
This is 400 real ground truth images  plus 1000 synthetic images for the particular GAN.

We use YOLOv5\cite{yolov5} as our object detection network for panicles.
We use the mean average precision (mAP) with Intersection over Union (IoU) from 0.5 to 0.95 (COCO mAP\cite{coco_dataset}) as the metric for detection, and Root Mean Squared Error (RMSE)\cite{powers_2011}, Mean Absolute Error (MAE)\cite{powers_2011} and Absolute Percent Error (MAPE)\cite{powers_2011} for counting:
\begin{equation}
    \text{Precision (P)} = \frac{\text{TP}}{\text{TP}+\text{FP}}
\end{equation}
\begin{equation}
    \text{Recall (R)} = \frac{\text{TP}}{\text{TP}+\text{FN}}
\end{equation}
\begin{equation}
    \label{eqa:ap}
    \text{AP} = \sum_{k}(R_{k}-R_{k-1})P_{k}
\end{equation}
\begin{equation}
    \text{MAPE} = \frac{1}{N} \sum_{\substack{i=1}}^{N}\frac{\big| e_i \big|}{C_i}
\end{equation}
\begin{equation}
  \label{eqa:mae}
    \text{MAE} = \frac{1}{N}\sum_{i=1}^{N}| e_i |
\end{equation}
\begin{equation}
  \label{eqa:rmse}
    \text{RMSE} = \sqrt{\frac{1}{N}\sum_{i=1}^{N} \big| e_i \big|^2}
\end{equation}
where $k$ refers to the $k$-th threshold for precision and recall.
$C_i$ is the ground truth count in the $i$-th image.
$e_i$ is the difference between $C_i$ and the estimated count.
$N$ is the number of image samples.
In our case, the mAP is equal to AP because we only have one class.
We evaluate the performance with the rest of 100 real ground truth images not used for training.
\begin{table}[htb]
\centering
\begin{tabular}[c]{cccc}
\toprule
\textbf{Metric}  &  \textbf{Real Image}  &  \textbf{pix2pixHD}  &  \textbf{SPADE}  \\\midrule
mAP@[.5,.95]     &  $72.4$          &  $78.9$        &  \textbf{79}         \\
MAPE    &  $11.6$          &  \textbf{7.2}        &  $9.7$         \\
MAE     &  $8.0$          &  \textbf{4.6}        &  $5.5$         \\
RMSE    &  $9.6$          &  \textbf{5.6}        &  $6.5$         \\\bottomrule
\end{tabular}
\caption{Panicle detection and counting results. Each column represents a detection model trained on the corresponding dataset. "Real Image" is the dataset of 400 real images. "pix2pixHD" and "SPADE" are the datasets with 400 real images plus 1000 synthetic images from the corresponding GAN. Bold indicates best performance.}
\label{tab:testing}
\end{table}
The results are shown in \Cref{tab:testing}.
Compared to training only on 400 real ground truth images, both GAN-based data augmentation methods achieve better mAP.
For the counting metrics, the model trained on the augmented images (real plus synthetic) also performs better.
Despite the different image styles generated by pix2pixHD and SPADE, their data augmentation performances are similar.
The synthetic images of pix2pixHD have slightly better performance for the counting metrics due to the more realistic background.
We found that for the pix2pixHD the weights tend to diminish around 200 epochs and the result becomes a completely black image.
SPADE does not have this problem, possibly due to the added SPADE residual blocks in the generator.
We feel both methods are suitable for GAN-based data augmentation for high-resolution UAV images.

\section{Conclusion}
\label{sec:conclu}
In this paper, we presented the use of GAN generated synthetic images to augment the training data for panicle detection and counting. 
We examined two image-to-image translation GANs and showed that their use can improve the performance of panicle detection and counting.
We did not use the temporal information available in our real Sorghum UAV dataset during training due to the limitation of the network structures.
Future work includes developing multi-temporal methods that can generate synthetic plant images in a temporally consistent style.
This will also us to estimate phenotypic traits as the plant grows.
We will also examine our approach for estimating traits of other plant such as maize tassels.

\section{Acknowledgments}

We thank Professor Ayman Habib and the Digital Photogrammetry Research Group (DPRG) from the School of Civil Engineering at Purdue University for providing the images used in this paper. The work presented herein was funded in part by the Advanced Research Projects Agency-Energy (ARPA-E), U.S. Department of Energy, under Award Number DE-AR0001135.
The views and opinions of the authors expressed herein do not necessarily state or reflect those of the United States Government or any agency thereof.
Address all correspondence to Edward J. Delp, ace@ecn.purdue.edu

{\small
\bibliographystyle{ieee_fullname}
\bibliography{ref}
}

\end{document}